\documentclass[11pt,onecolumn]{article}

\usepackage[T1]{fontenc}
\usepackage{lmodern}
\usepackage{amsmath}
\usepackage{amssymb}
\usepackage{geometry}
\usepackage{graphicx}
\usepackage{tikz}
\usetikzlibrary{shapes,arrows,positioning,calc,decorations.pathreplacing}
\usepackage{pgfplots}
\pgfplotsset{compat=1.18}

\usepackage{booktabs}
\usepackage{array}
\usepackage{tabularx}
\usepackage{multirow}
\usepackage{makecell}

\usepackage{algorithm}
\usepackage{algpseudocode}

\usepackage{hyperref}
\usepackage{natbib}
\bibpunct{(}{)}{;}{a}{,}{,}

\usepackage{xcolor}

\title{%
Persona-Trained Monte Carlo: Estimating Market-Outcome Distributions via \\
Swarms of Persona-Conditioned Neural Policy Bots in a Limit Order Book \\
}
\author{Salavat Ishbulatov \\
Independent researcher \\
\texttt{salavat@doplan.ai}}
\date{}

\begin{document}

\maketitle

% === ABSTRACT ===

\begin{abstract}
We propose Persona-Trained Monte Carlo (PTMC): a method for estimating distributions of market-outcome statistics by repeatedly simulating limit-order-book interaction among swarms of persona-conditioned neural policy bots, with an outer Monte Carlo loop over draws from a learned trader-heterogeneity distribution $\mathcal{P}$. Each simulation run instantiates $K$ bots sharing one trained policy architecture $\pi_\phi$ but conditioned on heterogeneous, individually-sampled persona draws $(\theta^{(k)}, \rho^{(k)}) \sim \mathcal{P}$; bots interact in a continuous double auction, and the resulting price path is one Monte Carlo sample. Repeating this $N_{\text{runs}}$ times over independent persona-population draws yields an ensemble from which a target functional $F$ (e.g., maximum drawdown, tail index, crash probability) is estimated via $\hat{\mu}_N = \frac{1}{N_{\text{runs}}}\sum_i F(\text{path}_i)$. Randomness therefore enters through three channels---persona draws, within-run action sampling from $\pi_\phi$, and optional exogenous shocks---rather than solely through an exogenous price process as in classical Monte Carlo. We distinguish PTMC from four adjacent paradigms: classical Monte Carlo (geometric Brownian motion, randomness only in price), hand-coded agent-based models (fixed behavioral archetypes, no learned $\mathcal{P}$), single-agent reinforcement learning (one optimized policy, not an ensemble over heterogeneity), and large-language-model-based generative agents (language-driven reasoning rather than a compact, shared trained network).

To justify this design, we survey cross-disciplinary foundations---agent-based computational economics, market microstructure, behavioral finance and neuroeconomics, deep RL for trading, generative and LLM-based agents, news-driven trading, systemic risk, econophysics, game theory, and the mathematical machinery of stochastic processes and information theory---connecting each literature explicitly to a specific design choice in $\pi_\phi$, the training data, or the validation protocol. We formalize the PTMC estimator and its convergence properties, specify a candidate bot architecture and training objective, and propose a validation methodology including stylized-fact matching, microstructure- and agent-level checks, and historical stress-test comparison, with explicit head-to-head tests against a zero-intelligence baseline. The framework is proposed but not implemented. We contribute a formal estimator, a cross-disciplinary design justification, and a validation roadmap, without reporting new simulation or empirical findings. We address ethical, systemic-risk, and privacy considerations, and conclude with open research questions.
\end{abstract}

\textbf{Keywords:} agent-based computational finance; Monte Carlo simulation; market microstructure; persona-conditioned agents; limit order book; behavioral cloning; reinforcement learning; financial market simulation

\newpage

\section{Introduction}
\label{sec:intro}

This paper proposes \textbf{Persona-Trained Monte Carlo (PTMC)}: a method that estimates distributions of market-outcome statistics by repeatedly simulating limit-order-book interaction among swarms of persona-conditioned neural policy bots, with an outer Monte Carlo loop over draws from a learned trader-heterogeneity distribution $\mathcal{P}$. Where classical Monte Carlo in finance averages a payoff functional over random paths of an exogenously specified price process \cite{cox_option_1979}, PTMC averages a target functional over random paths generated \emph{endogenously} by a population of interacting agents whose heterogeneity is itself a random draw, not a fixed design choice. This is the paper's central methodological proposal, and the survey that follows exists to justify the design choices that go into it---not the other way around.

Financial market prices emerge from the interaction of thousands of autonomous traders with heterogeneous information, expectations, and behavioral biases, yet most quantitative finance rests on homogeneous-agent models: Black-Scholes assumes a representative investor with constant risk aversion \cite{black_scholes_1973}; CAPM assumes rational agents \cite{sharpe_capital_1964}; the efficient markets hypothesis assumes prices instantly reflect all available information \cite{fama_efficient_1970}. Real markets violate these assumptions systematically, exhibiting fat tails, volatility clustering, and long memory that rational-agent models struggle to reproduce \cite{cont_stylized_2001}. Agent-based computational models (ABMs) address this by letting heterogeneous traders interact endogenously, generating complex price dynamics from simple local rules \cite{arthur_asset_1997, lux_volatility_1999, hommes_heterogeneous_2006}, but most hand-code behavioral rules or fixed archetypes rather than learning trader heterogeneity from data \cite{kahneman_prospect_1979, shefrin_disposition_1985}.

PTMC's departure from existing ABM practice is specific: instead of a fixed population of hand-specified trader types, each simulation run samples $K$ bots that share one trained policy network $\pi_\phi$ but are individually conditioned on persona draws $(\theta^{(k)}, \rho^{(k)}) \sim \mathcal{P}$, where $\mathcal{P}$ is itself learned from real trader behavioral and demographic data rather than specified by the modeler. Repeating this sampling-and-simulation procedure across many independent persona-population draws---the outer Monte Carlo loop---produces an ensemble of simulated market paths from which a target functional (maximum drawdown, tail index, spread distribution, crash probability) can be estimated with a quantified Monte Carlo error, exactly as in classical Monte Carlo valuation, except that the randomness being integrated over is trader heterogeneity itself rather than an exogenous stochastic price process. Section~\ref{sec:framework} (Conceptual Overview) makes this estimator precise.

This positions PTMC against four adjacent paradigms that are easy to conflate with it but differ in a specific, statable way. Classical Monte Carlo (e.g., Black-Scholes path simulation) randomizes only the exogenous price process; trader behavior is not modeled at all. Hand-coded ABMs (Santa Fe, Lux-Marchesi) randomize over a population, but the behavioral archetypes and their mixture weights are fixed by the modeler, not learned from a data-fitted distribution $\mathcal{P}$. A single reinforcement-learning trading agent optimizes one policy against a reward signal; it is not an ensemble over a heterogeneity distribution and has no outer Monte Carlo loop. LLM-based generative multi-agent systems (e.g., TwinMarket \cite{yang_twinmarket_2025}) instantiate personas through natural-language prompting and role-play rather than through a compact, shared, trained neural network, trading interpretability and reproducibility for flexibility; we discuss this alternative agent representation in Section~\ref{sec:llm} but it is not the comparison this paper centers. Bayesian neural networks, Monte Carlo dropout, and other single-model uncertainty-quantification techniques (Section~\ref{sec:uq-comparison}) randomize over one model's parameters or activations to express predictive uncertainty about a single forecasting task; PTMC instead randomizes over a population of interacting decision-makers whose collective behavior generates the outcome being measured, which is a different object of inference entirely.

Persona-conditioned bot agents---autonomous neural-network models trained on granular behavioral and demographic data to represent plausible market participants---are the building block that makes the outer Monte Carlo loop meaningful: without a learned $\mathcal{P}$ grounded in real heterogeneity, sampling personas would be no more informative than sampling from an arbitrary prior. Each bot ingests market state (prices, order-book depth, news) and personal state (demographics, behavioral profile) and outputs a trading decision via the shared learned policy $\pi_\phi$, operating in a continuous double auction limit order book exactly as human participants do, including access to an explicit external-information channel for financial news, macroeconomic indicators, and index-level trend signals.

The paper is structured as follows. Section~\ref{sec:related} surveys the cross-disciplinary foundations that justify PTMC's design choices---agent-based economics, market microstructure, behavioral finance, deep RL for trading, generative and LLM-based agents, news-driven trading, systemic risk, econophysics, game theory, neuroeconomics, and mathematical foundations---closing with a synthesis mapping literature to design element. Section~\ref{sec:framework} formalizes PTMC: the Monte Carlo estimator, bot architecture, training objective, and market mechanism. Section~\ref{sec:validation} proposes a validation methodology, including head-to-head comparison against a zero-intelligence baseline. Section~\ref{sec:discussion} addresses limitations, systemic risk, and ethics. Section~\ref{sec:conclusion} outlines open research questions.

We are explicit throughout about what is proposed versus proven: this paper formalizes an estimator and a research agenda, grounded in a cross-disciplinary survey, but reports no implementation and no empirical results. A reader finishing this introduction should know what PTMC estimates, how its randomness differs from classical Monte Carlo, hand-coded ABMs, single RL agents, and LLM agents, and that everything past this point is a falsifiable proposal, not a finding.

\section{Background and Related Work}
\label{sec:related}

This section surveys literatures that motivate PTMC's design---agent-based economics, market microstructure, behavioral finance, machine learning for trading, generative and LLM agents, news-driven trading, systemic risk, econophysics, game theory, mathematical foundations, and validation practice---organized by conceptual dependency, not chronology. Each subsection states what it contributes to $\pi_\phi$, training data, the external-information channel (Section~\ref{sec:framework}), or the validation protocol (Section~\ref{sec:validation}); unresolved disputes in the cited work are noted explicitly, and PTMC inherits rather than settles them.

\subsection{Positioning Relative to Existing Surveys}
\label{sec:positioning}

Several of the subfields surveyed below already have dedicated, mature survey literatures of their own, and a reasonable question for any reader specializing in one of these subfields is what this paper adds beyond reading the relevant dedicated survey directly. We address this explicitly rather than leaving it implicit.

On deep reinforcement learning for algorithmic trading, Pippas, Ludvig, and Turkay's recent \emph{ACM Computing Surveys} article \citep{pippas_evolution_2025} systematically reviews 167 publications on RL applied to trading, portfolio management, and market-making, providing far deeper coverage of RL algorithm variants, reward formulations, and empirical trading performance than Section~\ref{sec:rl} of this paper attempts. On agent-based computational economics and finance, Axtell and Farmer's synthesis \citep{farmer_axtell_2025} and the foundational \emph{Handbook of Computational Economics} \citep{tesfatsion_agent-based_2006} provide comprehensive treatments of ABM methodology, calibration practice, and policy applications across the full span of economics, not only finance, with far greater methodological depth than Section~\ref{sec:abe} of this paper. On LLM-based generative agents, Mou et al.'s survey of social simulation driven by LLM-based agents \citep{mou_individual_2024} comprehensively categorizes individual-, scenario-, and society-level LLM simulation architectures across social science applications broadly, again with more depth than Section~\ref{sec:llm} of this paper devotes to the topic.

We do not out-survey these literatures on their own terms; readers seeking depth should consult them directly. We instead connect behavioral finance, microstructure, mathematical foundations, and external information (Sections~\ref{sec:bf}, \ref{sec:mm}, \ref{sec:math}, and~\ref{sec:news}) into one PTMC target architecture (Table~\ref{tab:design-mapping})---breadth of synthesis, not depth within any single subfield.

This is a selective, narrative review organized by conceptual dependency rather than a systematic literature map: its purpose is to justify PTMC's design choices (Table~\ref{tab:design-mapping}), not to provide exhaustive, protocol-driven coverage of any one subfield.

\subsection{Classical Financial Theory and Its Limitations}
\label{sec:classical}

Modern financial theory rests on mid-20th-century foundations: mean-variance portfolio selection \cite{markowitz_portfolio_1952}, the Capital Asset Pricing Model \cite{sharpe_capital_1964}, Black-Scholes-Merton option pricing under geometric Brownian motion \cite{black_scholes_1973, merton_rational_1973, cox_option_1979}, and the general-equilibrium theory \cite{arrow_general_1954} establishing conditions (complete markets, perfect information, rational agents) under which decentralized trading achieves Pareto efficiency. The efficient markets hypothesis (EMH) \cite{fama_efficient_1970} claims prices fully incorporate available information. Each assumes a representative, rational agent and a single specified stochastic process---assumptions empirical evidence violates systematically.

Fama and French (1993) documented size and book-to-market effects beyond market beta \cite{fama_three-factor_1993}; De Bondt and Thaler (1985) found long-horizon mean reversion \cite{de_bondt_anomaly_1985} while Jegadeesh and Titman (1993) found intermediate-horizon momentum \cite{jegadeesh_returns_1993}; Carhart (1997) attributed fund performance persistence to momentum exposure, not skill \cite{carhart_mutual_1997}. Shiller (1981) showed prices are far more volatile than rational dividend-forecast models predict \cite{shiller_speculative_1981}. Real investors also fail to rebalance optimally as lifetime portfolio theory \cite{merton_lifetime_1969} prescribes, exhibiting loss aversion and anchoring instead. Returns more broadly show fat tails, volatility clustering, and long memory absent from random walks.

The disagreement between EMH/rational-expectations theory and behavioral finance (Section~\ref{sec:bf}) is not merely a matter of degree---it is a live methodological dispute about what counts as disconfirming evidence. Fama's defenders argue that anomalies (momentum, value, size effects) are themselves compensation for unmodeled risk factors, so that violations of the simplest EMH specification are not violations of market efficiency per se, only of a particular asset-pricing model (the ``joint hypothesis problem'': any test of efficiency is simultaneously a test of the pricing model used to define normal returns) \cite{fama_efficient_1970, fama_three-factor_1993}. Behavioral economists counter that anomalies persist out-of-sample and are predictable in direction (e.g., momentum reverses at long horizons \cite{de_bondt_anomaly_1985,jegadeesh_returns_1993}), which is difficult to reconcile with any risk-based explanation that does not itself require ad hoc, anomaly-specific risk factors. Neither side has produced a single nested model that the other accepts as decisive; the dispute over whether prices are ``right but for measurable risk'' or ``systematically biased by behavioral frictions'' remains unresolved in the literature this paper draws on, and persona-trained Monte Carlo inherits this unresolved question rather than settling it: if bot behavior reproduces an anomaly, that is consistent with both a risk-based and a behavioral account of why the anomaly exists in real markets.

Classical theory supplies the null hypothesis against which persona-bot dynamics are judged (the GBM/i.i.d.\ degenerate case) and the anomalies above are the target regularities; see Table~\ref{tab:design-mapping} for the full literature-to-design mapping.

\subsection{Agent-Based Computational Economics and Financial Markets}
\label{sec:abe}

Agent-based computational economics (ACE) studies economic systems as evolving networks of interacting autonomous agents \cite{tesfatsion_agent-based_2006}, replacing the equilibrium/rationality assumption with heterogeneous, boundedly rational, adaptively learning agents whose repeated interaction generates emergent dynamics. The conceptual roots trace to Vernon Smith's laboratory experiments showing that a simple trading institution (the continuous double auction) can achieve allocative efficiency without agents possessing global information or rational expectations \cite{smith_experimental_1962}---market structure and simple behavioral rules can induce macro order without agent rationality.

The modern era began with the Santa Fe Artificial Stock Market, where Arthur, Holland, LeBaron, Palmer, and Tayler populated a computational market with genetic-algorithm-evolved forecasting rules and generated endogenous bubbles, crashes, and fat-tailed returns absent from rational models \cite{arthur_asset_1997, palmer_artificial_1994}. Contemporaneously, Lux and Marchesi's stochastic model of fundamentalist/noise-trader switching reproduced volatility clustering and power-law returns \cite{lux_volatility_1999}, and Cont and Bouchaud showed herding-driven information cascades can amplify volatility absent any fundamental shock \cite{cont_bouchaud_2000}---establishing that multi-state, multi-trader ABMs reproduce stylized facts. Brock and Hommes formalized adaptive-belief switching between competing forecast models (trend-following vs.\ fundamental), producing bifurcations and time-varying equilibria \cite{brock_hommes_1998}, later unified by Hommes to explain persistent deviations from rational-expectations equilibrium \cite{hommes_heterogeneous_2006}. Levy, Levy, and Solomon's microscopic simulation approach tracked individual investor wealth and endogenous market participation, showing wealth effects and inequality drive macro price dynamics \cite{levy_levy_solomon_2000}, bridging behavioral finance and ABM---and constituting the closest conceptual precursor to PTMC's outer draw of a heterogeneous trader population, though their heterogeneity is parametric rather than learned from trader-level data.

By 2006 the field was mature enough for Tesfatsion and Judd's two-volume \textit{Handbook of Computational Economics} \cite{tesfatsion_agent-based_2006}, and Samanidou et al.\ surveyed ABM applications to microstructure, volatility, and bubbles \cite{samanidou_agent-based_2007}.

Table~\ref{tab:abm-comparison} summarizes the properties of the principal agent-based market models surveyed above, drawing only on properties already established in the cited primary sources.

\begin{table}[htb]
\centering
\small
\begin{tabular}{p{2.6cm}p{2.5cm}p{2.9cm}p{1.7cm}p{1.7cm}p{3.2cm}}
\toprule
\textbf{Model} & \textbf{Data Requirements} & \textbf{Stylized Facts Reproduced} & \textbf{Comp. Cost} & \textbf{Interpre-}\newline\textbf{tability} & \textbf{Key Limitation} \\
\midrule
Santa Fe ASM \citep{arthur_asset_1997,palmer_artificial_1994} & None (GA-evolved rules, no real trader data) & Bubbles, crashes, fat tails & Medium & High (explicit forecasting rules) & Heterogeneity is GA-evolved, not empirically grounded in real trader behavior \\
\addlinespace
Lux-Marchesi \citep{lux_volatility_1999} & None (hand-specified switching rule) & Volatility clustering, power-law returns & Low & High & Only two trader types (fundamentalist/noise); switching mechanism is stylized, not estimated from data \\
\addlinespace
Cont-Bouchaud \citep{cont_bouchaud_2000} & None (random-graph clustering parameter) & Herding-driven fat tails & Low & Medium & Cluster/herd formation is exogenously parameterized, not learned \\
\addlinespace
Heterogeneous-Agent Model (Brock-Hommes) \citep{brock_hommes_1998,hommes_heterogeneous_2006} & None (belief-type fitness weights) & Bifurcations, momentum, mean reversion & Medium & Medium (nonlinear dynamics less transparent) & Small number of competing forecast rules; not validated against individual trader heterogeneity \\
\addlinespace
Gode-Sunder zero-intelligence \citep{gode_allocative_1993} & None (budget constraint only) & Allocative efficiency only; \emph{not} fat tails/clustering & Very low & Very high & Reproduces efficiency but not the time-series stylized facts central to Section~\ref{sec:cal} \\
\addlinespace
\textbf{Persona-trained neural-bot (proposed)} & \textbf{High} (real trader transaction/demographic data) & \textbf{Targeted: all of the above plus agent-level fidelity (untested)} & \textbf{Medium-High} & \textbf{Medium} & \textbf{No implementation yet exists; equifinality (Section~\ref{sec:cal}) means stylized-fact matching alone would not confirm mechanism realism} \\
\bottomrule
\end{tabular}
\caption{Comparison of principal agent-based stock-market models against the proposed persona-trained framework, restricted to properties established in the cited primary sources. No row reports results from an implementation of the proposed framework, since none exists; the final row describes design targets, not measured outcomes.}
\label{tab:abm-comparison}
\end{table}

Gode and Sunder's zero-intelligence result \citep{gode_allocative_1993} is a direct, unresolved objection to the core premise of this paper, not a methodological footnote. Their experiment showed that traders with no memory, no learning, and no strategic reasoning---constrained only by a budget---achieve allocative efficiency in a double auction indistinguishable from that of experienced human traders. If a near-zero-intelligence population is sufficient for the institution (the CDA) to produce efficient, even realistic-looking, aggregate outcomes, then the entire research program of building psychologically rich, demographically-conditioned bot personas is at risk of solving a problem the matching mechanism already solves on its own. The ABE literature has not settled this objection: subsequent work shows zero-intelligence models reproduce \emph{allocative efficiency} but do not reproduce the fat tails, volatility clustering, or long-memory stylized facts documented by Cont \citep{cont_stylized_2001} that Lux-Marchesi-style heterogeneous-agent models do reproduce \citep{lux_volatility_1999}---suggesting that efficiency and stylized-fact realism are separable properties, with zero intelligence sufficient for the former but apparently not the latter. This paper's working position, stated explicitly here rather than left implicit, is that behavioral realism is justified by the stylized-fact target, not the efficiency target; persona-trained Monte Carlo must therefore be evaluated on whether richer behavioral modeling earns its computational and data cost specifically on dimensions where zero-intelligence agents are known to fail, a claim that itself requires the head-to-head comparison we did not find directly addressed in the cited literature and which Section~\ref{sec:validation} is designed to test.

Altogether, agent-based models of financial markets decouple price dynamics from rational-expectations equilibrium, allowing heterogeneous agents with bounded rationality and adaptive learning to generate emergent phenomena (volatility clustering, bubbles, crashes) that match empirical stylized facts. The framework is now mature and widely used in academic research and policy analysis.

ACE is PTMC's direct methodological ancestor, differing in how heterogeneity is obtained (learned vs.\ hand-specified); see Table~\ref{tab:design-mapping}.

\subsection{Market Microstructure and Trading Mechanisms}
\label{sec:mm}

Kyle's canonical model of informed trading has a strategic insider competing with noise traders against a risk-neutral market maker who infers information from order flow \cite{kyle_continuous_1985, kyle_information_1989}---prices aggregate dispersed information through the order-matching process itself. Glosten and Milgrom instead model the bid-ask spread as compensation for adverse selection \cite{glosten_bid_1985}, formalized by Easley and O'Hara's Probability of Informed Trading measure \cite{easley_informed_1987}. Green showed no mechanism can simultaneously achieve allocative efficiency, individual rationality, and incentive compatibility under private information \cite{green_information_1986}, and Grossman and Stiglitz showed markets cannot be perfectly informationally efficient once information acquisition is costly \cite{grossman_noise_1980}.

Smith's continuous double auction achieves high allocative efficiency even with boundedly rational traders \cite{smith_experimental_1962}; Gode and Sunder's zero-intelligence traders pushed this further, achieving near-perfect efficiency with no foresight at all \cite{gode_allocative_1993}---efficiency is often a property of the institution, not the agents. Modern electronic markets implement this via the limit order book, surveyed by Parlour and Seppi \cite{parlour_limit_2008}, with Hasbrouck \cite{hasbrouck_asset_2007} and Harris \cite{harris_microstructure_2003} providing comprehensive empirical and theoretical treatments of order flow, spreads, depth, and price impact. Chiarella, He, Shi, and Wei's behavioral LOB model calibrates fundamentalist, chartist, and noise-trader agent behaviors directly from data within a limit-order-book mechanism \citep{chiarella_behavioural_2017}, and Gould et al.'s survey of limit order books \citep{gould_limit_2013} catalogs spreads, depth, price impact, and trade-size distributions as the microstructure benchmarks against which any LOB-based simulation should be assessed; both are direct methodological precedents for PTMC's market mechanism and validation protocol.

Cont's synthesis of empirical stylized facts---fat tails, volatility clustering, long memory, the leverage effect, and the absence of raw-return autocorrelation alongside positive autocorrelation in absolute/squared returns \cite{cont_stylized_2001}---is the central validation target for any ABM, and Madhavan's survey frames the resulting tension between information efficiency and allocative efficiency \cite{madhavan_microstructure_2000}.

Kyle's framework \citep{kyle_continuous_1985} and the noise-trading/behavioral-finance literature surveyed in Section~\ref{sec:bf} make structurally opposed predictions about what prices \emph{are}. In Kyle's model, prices are a noisy but ultimately accurate aggregator of private information: order flow reveals the insider's signal, and the market maker's pricing rule converges toward fundamental value as informed trading accumulates. DeLong, Shleifer, Summers, and Waldmann's noise-trader model \citep{delong_noise_1990} and the broader behavioral-finance evidence (excess volatility \citep{shiller_speculative_1981}, persistent anomalies \citep{fama_three-factor_1993}) instead treat price as frequently and persistently distorted by uninformed demand that rational arbitrageurs cannot fully correct because arbitrage itself is risky and capital-constrained. These are not two descriptions of the same mechanism at different time horizons; they imply different things about whether order flow should be trusted as an information signal at all. Grossman and Stiglitz's result \citep{grossman_noise_1980} that markets cannot be perfectly informationally efficient sits awkwardly between the two views---it requires noise traders for informed traders to profit, but does not by itself establish whether the resulting prices are ``close to'' fundamental value (Kyle-consistent) or systematically biased away from it (noise-trader-consistent) over policy-relevant horizons. A persona-bot simulation that gives bots access to order-flow features as in Kyle's tradition, while also training those same bots on behaviorally biased real trader data, is implicitly taking a position in this dispute (that both information aggregation and persistent distortion coexist) without the literature offering a settled account of how the two interact quantitatively.

Market microstructure supplies the literal CDA/LOB mechanism and the order-flow features in $s_t$; see Table~\ref{tab:design-mapping}.

\subsection{Behavioral Finance and Investor Psychology}
\label{sec:bf}

Kahneman and Tversky's prospect theory \cite{kahneman_prospect_1979} showed experimentally that people exhibit reference-dependent preferences (losses loom larger than gains), diminishing sensitivity, and probability weighting---violating expected utility theory; their earlier work on judgment heuristics (availability, representativeness, anchoring) documented the systematic errors these shortcuts produce \cite{tversky_judgment_1974}.

In trading specifically: Shefrin and Statman's disposition effect (selling winners, holding losers) \cite{shefrin_disposition_1985} has been extensively replicated \cite{odean_overconfidence_1998}; DeLong, Shleifer, Summers, and Waldmann's noise-trader model formalized how correlated mispricing by psychologically-driven traders can persist when rational arbitrage is capital- or hedging-constrained \cite{delong_noise_1990}; and Barber and Odean showed active traders underperform buy-and-hold investors, attributed to overconfidence \cite{barber_trading_2002}, while Lakonishok, Shleifer, and Vishny showed representativeness-driven extrapolation explains both momentum and reversal \cite{lakonishok_contrarian_1994}. Barberis and Thaler \cite{barberis_survey_2003}, Shiller's ``irrational exuberance'' \cite{shiller_irrational_2000}, Statman \cite{statman_behavioral_2017}, and Lo's adaptive markets hypothesis \cite{lo_market_2012} synthesize this evidence into the broader claim that market efficiency is contextual and time-varying with agent composition.

At the market level, Kirman's contagion model shows herding and speculative bubbles can emerge from simple individual-level switching dynamics with no external shock \cite{kirman_heterogeneous_1991}; Blume and Easley showed that market selection dynamics favor traders whose strategies better match the true data-generating process, so markets with heterogeneous beliefs can take a long time to converge to rational-expectations outcomes \cite{blume_wealth_1993}; and Daniel, Hirshleifer, and Subrahmanyam's model of overreaction and underreaction \cite{daniel_investor_1998} together with Daniel and Moskowitz's documentation of momentum crashes \cite{daniel_systematic_2016} further document patterns difficult to reconcile with rational pricing.

Even within the literature that documents herding as a robust phenomenon, there is no consensus on whether herding is a psychological bias or a rational response to incentives and information frictions. Scharfstein and Stein \citep{scharfstein_herd_1990} model herding as \emph{reputational}: fund managers rationally mimic peers to protect their career prospects, not because they misperceive information, and this herding is privately optimal even though it is socially inefficient. Bikhchandani and Sharma's review \citep{bikhchandani_herd_2000} formalizes a parallel ``informational cascade'' channel in which agents rationally discount their own private signal in favor of observed public actions---again, fully consistent with Bayesian rationality. Kirman's contagion model \citep{kirman_heterogeneous_1991}, by contrast, treats herding as a process more akin to epidemiological spread of belief, with no requirement that any individual agent's switching decision be optimal in a reputational or informational sense. The training data described in Training Data Sources (Section~\ref{sec:framework}) cannot, by construction, distinguish between these mechanisms merely by observing that a trader's action correlates with others' recent actions; a persona-bot trained to reproduce the correlation will be agnostic about which of these competing causal stories is responsible for it, which limits the interpretive claims that can be drawn from any herding behavior the simulation reproduces.

Behavioral finance is the direct source of the behavioral profile $\rho$ and loss-aversion coefficient $\lambda$; see Table~\ref{tab:design-mapping}.

Behavioral constructs above motivate explicit $\lambda$ and $\rho$; neuroeconomic evidence further suggests treating gain/loss and conformity as separable computations \citep{kuhnen_neural_2005, demartino_amygdala_2010, klucharev_downregulation_2011}, motivating distinct encoder/value branches in $\pi_\phi$ (design prior only; behavioral cloning remains primary).

\subsection{Deep Reinforcement Learning for Algorithmic Trading}
\label{sec:rl}

Reinforcement learning (RL) learns a state-to-action policy by trial and error, maximizing cumulative reward; deep RL parameterizes this policy with neural networks, enabling learning in high-dimensional state spaces. The foundations---Bellman's dynamic programming \cite{bellman_dynamic_1957}, Watkins's Q-learning \cite{watkins_q-learning_1992}, Sutton and Barto's policy-gradient methods \cite{sutton_policy_2000}, and Konda and Tsitsiklis's actor-critic methods \cite{konda_actor-critic_2003}---were scaled to high-dimensional inputs by Mnih et al.'s DQN on Atari \cite{mnih_playing_2013} and Silver et al.'s AlphaGo \cite{silver_mastering_2016}.

Applied to trading, Millea's survey \cite{almahdi_deep_2021} identifies non-stationary environments, sparse episodic rewards, transaction costs, and interpretability as the field's persistent challenges across DQN, policy-gradient (A3C, PPO), and actor-critic approaches to portfolio optimization, market-making, and trading. Continuous-control methods (DDPG-family) address optimal execution \cite{almgren_optimal_2001}; Lalor and Swishchuk address non-stationarity by modeling price dynamics with semi-Markov and Hawkes jump-diffusion processes rather than a standard Markov assumption \cite{lalor_reinforcement_2024}; and Vadori et al.\ study coordination and competitive dynamics in a multi-agent liquidity-provider/liquidity-taker RL setting for OTC markets \cite{vadori_otc_2024}---closer to PTMC's setting than single-agent RL, but markedly harder to analyze since the state space now includes other agents' evolving strategies.

There is an unresolved tradeoff, not a solved engineering problem, between reward-maximizing RL policies and behavioral realism. Pippas, Ludvig, and Turkay's survey of 167 RL-for-finance publications \citep{pippas_evolution_2025} documents that the field's dominant evaluation criterion remains risk-adjusted return (Sharpe ratio, cumulative P\&L), not similarity to real human trading behavior; an RL agent that discovers a profitable strategy no human trader uses is, by the field's own standard metrics, a successful result. This is precisely the opposite optimization target from what a persona-bot needs: a policy trained by pure reward maximization has no pressure to resemble the demographic- and psychology-conditioned behavior documented in Section~\ref{sec:bf}. Behavioral cloning, this paper's proposed corrective, solves the opposite problem (it reproduces observed actions faithfully) but inherits its own well-known failure mode---compounding distributional-shift error once the bot's own actions perturb the market away from the state distribution in the training data, a problem behavioral cloning cannot self-correct because it has no reward signal telling it when its policy has drifted. The hybrid loss proposed in Section~\ref{sec:framework} (Algorithms~\ref{alg:bc}--\ref{alg:hybrid}) is a candidate resolution, not a demonstrated one: nothing in the cited literature establishes the weighting $(\alpha, \beta, \gamma)$ at which behavioral fidelity and profit-seeking adaptation can be jointly satisfied, and it is entirely possible that any nonzero reward weight $\beta$ erodes behavioral realism in proportion to how profitable the discovered deviation is.

The hybrid behavioral-cloning/IRL objective of Section~\ref{sec:framework} is a direct synthesis of this literature's reward-vs.-realism tension; see Table~\ref{tab:design-mapping}.

\subsection{Generative Models, GANs, and Large Language Model-Based Agents}
\label{sec:llm}

Recent advances in large language models (LLMs) and generative AI have opened new possibilities for agent-based simulation. Rather than hand-coding agent behaviors, one can train a generative model on human behavioral data and use the trained model to populate a simulation.

\textbf{Generative models and their foundations:}

Generative modeling provides the technical substrate for both PTMC's bot policies and the LLM-agent alternative discussed below. Variational autoencoders \cite{kingma_auto-encoding_2014} and GANs \cite{goodfellow_generative_2014} learn to produce realistic samples from high-dimensional data and have been applied directly to financial time series and order-flow synthesis \cite{tang_survey_2021, rizzato_generative_2023, wiese_quant_2020}; Hultin et al.'s generative recurrent-network approach learns realistic limit-order-book dynamics directly from tick data \citep{hultin_generative_2023}, the closest learned-generative precedent to PTMC's bot-level policy among purely data-driven approaches. On the language side, embedding techniques \cite{mikolov_word2vec_2013, pennington_glove_2014} and the Transformer architecture \cite{vaswani_attention_2017, devlin_bert_2019, radford_language_2019, brown_language_2020, bengio_scaling_2021} underlie the LLM-based generative-agent paradigm: Park et al.'s Generative Agents \cite{park_generative_2023} showed that LLMs equipped with persona context, episodic memory, reflection, planning, and inter-agent interaction can produce convincing emergent social behavior (e.g., autonomously organized social events) without explicit programming, and subsequent work has surveyed agentic LLM systems more broadly \cite{plaat_agentic_2025} or multi-agent geosimulation methodologies spanning classical ABMs to LLM-based agents specifically \cite{padilla_survey_2025}, or built general-purpose multi-agent LLM social-simulation platforms directly on this premise \cite{gensim_2024}. The persistent open challenges for this paradigm---calibration against real behavior, inference cost at simulation scale, stochastic non-reproducibility, interpretability, and inherited training-data bias---are precisely what motivate the compact, narrowly-trained alternative pursued here.

\textbf{Alternative agent representation: LLM-based generative agents.} A structurally different way to instantiate persona-conditioned agents has emerged very recently: rather than a compact trained network, condition a large language model on a persona via natural-language prompting and role-play. Yang et al.'s TwinMarket \citep{yang_twinmarket_2025} simulates up to 1,000 individually-profiled LLM traders and reproduces fat-tailed returns, volatility clustering, and information-driven crashes; Yu et al.'s FinCon \citep{yu_fincon_2024} uses verbal reinforcement in place of gradient-based policy updates. Mou et al.'s survey of LLM-driven social simulation \citep{mou_individual_2024} catalogs this broader paradigm across individual-, scenario-, and society-level architectures; relative to a compact trained policy, the apparent trade-off is that LLM agents generalize to novel scenarios without retraining, at the likely cost of higher inference expense, weaker reproducibility under stochastic decoding, and reduced interpretability---properties this paper's framework is designed around rather than a finding directly reported in any single cited source. These systems are direct existence proofs that persona-conditioned agent populations can reproduce realistic stylized facts, but at a different point in the design space (full LLM reasoning rather than a shared trained network), and at scales (tens of agents, ten indices) well below what a market-wide PTMC simulation targets; we treat this as a parallel research direction rather than the comparison class for PTMC, which is detailed in Table~\ref{tab:paradigm-comparison}.

On balance, LLMs and generative models enable rich behavioral simulation but at high inference cost and low interpretability; compact, narrowly-trained neural networks offer interpretability and efficiency at the cost of less flexible persona representation. PTMC adopts the latter.

Table~\ref{tab:paradigm-comparison} extends the comparison across the full space of trading-agent paradigms surveyed in this Background section, including the equifinality risk identified in Section~\ref{sec:cal} as an explicit evaluation dimension.

\begin{table}[htb]
\centering
\footnotesize
\begin{tabular}{p{2.3cm}p{1.9cm}p{1.5cm}p{1.7cm}p{1.5cm}p{1.9cm}p{2.3cm}}
\toprule
\textbf{Paradigm} & \textbf{Behav.\ Realism} & \textbf{Data Needs} & \textbf{Interpret.} & \textbf{Comp.\ Cost} & \textbf{Validation Maturity} & \textbf{Equifinality Risk} \\
\midrule
Rule-based ABM (Lux-Marchesi, Santa Fe) & Low-Medium & Low & High & Low & Mature (decades of calibration work) & High (Section~\ref{sec:cal}) \\
\addlinespace
Zero-intelligence agents & Low & Very low & Very high & Very low & Mature for efficiency claims only & High for any stylized-fact claim \\
\addlinespace
Deep RL agents & Medium (reward-driven, not behavior-matched) & High (simulation episodes) & Low & High (training) & Emerging; \citep{pippas_evolution_2025} & Medium-High \\
\addlinespace
LLM-based generative agents & High (claimed) & Very high (pretraining + prompts) & Low & Very high (inference) & Early; few quantitative benchmarks \citep{mou_individual_2024} & Unknown---too new to assess \\
\addlinespace
\textbf{Persona-bots (proposed)} & \textbf{High (targeted, untested)} & \textbf{High} & \textbf{Medium} & \textbf{Medium} & \textbf{None---no implementation exists} & \textbf{High, by the same logic as all rows above} \\
\bottomrule
\end{tabular}
\caption{Comparison of trading-agent paradigms across behavioral realism, resource requirements, and validation maturity. The bottom row describes design targets for the proposed framework, not measured properties of an implementation. The equifinality risk column reflects the methodological point developed in Section~\ref{sec:cal}: every paradigm in this table, including the proposed one, is vulnerable to the objection that matching aggregate statistics is weak evidence for mechanism correctness.}
\label{tab:paradigm-comparison}
\end{table}

\subsection{News-Driven and Event-Driven Trading, and Financial Sentiment Analysis}
\label{sec:news}

Real market participants do not trade on price history and order flow alone: they read news, react to earnings announcements, macroeconomic releases, and geopolitical events, and update beliefs in response to qualitative information. Any realistic persona-bot architecture must incorporate this channel; the present framework does so via the external information channel of Section~\ref{sec:framework}.

Fedyk's natural experiment on front-page versus inside-page placement of otherwise identical news shows that attention allocation, not just content, causally drives trading volume and price response \citep{fedyk_frontpage_2024}, motivating salience-weighted news inputs; Hirshleifer and Sheng show macro- and firm-level news interact rather than substitute, motivating separate macro and firm-level input streams rather than one scalar sentiment score \citep{hirshleifer_macro_2022}. Araci's FinBERT \citep{araci_finbert_2019} is the domain-adapted transformer underlying the text-embedding approach used here in place of hand-engineered sentiment scores. Glasserman, Mamaysky, and Qin's entropy-based novelty measure---how unusual a news item is relative to recent news---predicts subsequent returns better than sentiment or uncertainty indices alone \citep{glasserman_new_2023}, suggesting bots should track novelty, not just polarity (we return to this information-theoretic framing in Section~\ref{sec:math}). TwinMarket \citep{yang_twinmarket_2025} and FinCon \citep{yu_fincon_2024} provide empirical support that an external news channel is load-bearing: the former reproduces information-driven crashes at the market level, the latter improves trading performance (Section~\ref{sec:llm}).

McLean and Pontiff \citep{mclean_does_2016} document that across 97 return-predictive characteristics, average returns fell 26\% out-of-sample and 58\% post-publication, consistent with sophisticated capital trading against published anomalies. This does not undermine news and novelty as bot input features---bots are meant to reproduce realistic trader \emph{reaction} to news, not to exploit price predictability---but validation exercises should not assume recent news-to-return relationships are stationary.

This literature motivates the External Information Channel of Section~\ref{sec:framework}; see Table~\ref{tab:design-mapping}.

\subsection{Network Dynamics, Systemic Risk, and Financial Contagion}
\label{sec:networks}

Real financial markets are interconnected ecosystems where institutional linkages create channels for shocks to propagate \citep{jackson_systemic_2021}. Foundational network science---small-world structure \cite{watts_collective_1998}, scale-free degree distributions vulnerable to targeted (not random) failure \cite{albert_small_1999}, and the broader synthesis of power laws and community structure \cite{newman_power_2005}---underlies the financial-contagion literature: Allen and Gale showed bank failures can propagate through interbank lending networks absent any fundamental shock \cite{allen_financial_2000}; Haldane and May showed financial networks exhibit phase transitions where small shocks trigger cascades \cite{haldane_systemic_2011}; Acemoglu, Ozdaglar, and Tahbaz-Salehi showed denser connectivity can paradoxically increase systemic risk under large shocks via the same ``network topology risk'' \cite{acemoglu_systemic_2015}; and Battiston et al.\ showed leverage interdependencies induce correlated defaults \cite{battiston_systemic_2012}, with empirical confirmation in real interbank networks \cite{bargigli_interbank_2016, glasserman_contagion_2015, haldane_economics_2009}.

The Flash Crash of May 6, 2010 (Dow down $\approx$9\%/998.5 points within minutes before recovering most losses) is the canonical systemic-event case study \cite{sec_flash_2010}; Kirilenko et al.'s audit-trail analysis found HFTs did not withdraw but rapidly passed accumulating inventory among themselves (a ``hot potato'' effect), producing functionally the same outcome as a liquidity withdrawal---unbuffered order-flow imbalance and excess price movement \cite{kirilenko_flash_2017}.

The network-contagion literature is internally divided on whether greater interconnectedness makes financial systems more or less fragile, and the answer the literature gives is explicitly non-monotonic rather than settled in either direction. Acemoglu, Ozdaglar, and Tahbaz-Salehi \citep{acemoglu_systemic_2015} show that for small shocks, denser interbank connectivity \emph{improves} risk-sharing and stability (the classical diversification argument), while for sufficiently large shocks the same dense connectivity \emph{amplifies} contagion, because the same channels that absorb small losses transmit large ones; the network-topology effect therefore reverses sign as shock size crosses a threshold that depends on parameters the model does not pin down empirically. Haldane and May's phase-transition framing \citep{haldane_systemic_2011} treats this reversal as a feature of complex systems generally, but neither paper offers a way to predict, ex ante, where the critical shock-size threshold lies for a real, given network. For a persona-trained simulation, this matters directly: if cascade dynamics are meant to emerge endogenously from correlated bot responses (as the Connection paragraph below argues), the simulation's realism with respect to systemic risk depends on getting bot-population correlation structure right at exactly the regime (the size and correlation of the shock) where the theoretical literature itself disagrees about whether connectivity helps or hurts---a regime-dependence the validation protocol of Section~\ref{sec:validation} would need to test for explicitly rather than assume away.

Altogether, financial markets are networks where contagion and systemic risk depend critically on network structure. Network topology, leverage, and trading strategy correlation create amplification mechanisms. Persona-trained simulations should include multiple agent types (retail, institutional, high-frequency traders) with realistic correlation structures to capture systemic-risk scenarios.

This network literature motivates instantiating multiple bot subpopulations with distinct $(\theta,\rho)$ distributions so cascade dynamics can emerge endogenously; see Table~\ref{tab:design-mapping}.

\subsection{Econophysics, Complexity, and Emergence}
\label{sec:econ-physics}

Physics and complexity science contribute several transferable insights. Bak's self-organized criticality shows complex systems naturally evolve toward a critical state where small perturbations trigger power-law-distributed cascades \cite{bak_self-organized_1987}, echoed in Kauffman's work on phase transitions and the ``edge of chaos'' in adaptive systems \cite{kauffman_self-organized_1990, kauffman_adaptive_1995}. Mandelbrot's observation that price changes are scale-invariant/fractal \cite{mandelbrot_fractals_1963} produces fat tails and volatility clustering inconsistent with normal-distribution models, and Sornette argues crashes are manifestations of this extreme-event physics rather than purely exogenous shocks \cite{sornette_why_2003}. Complexity economics more broadly \cite{beinhocker_origin_2006, farmer_economics_2012, mainzer_emergence_2010, farmer_adaptive_2008} treats economies as far-from-equilibrium adaptive systems where emergent phenomena are incomprehensible from individual-level description alone. Sandroni's result that the most-accurate-belief agent comes to dominate a repeated market through profit-driven trading, not direct selection, supports an evolutionary rather than static view of market efficiency \cite{sandroni_efficient_2000}.

Econophysics's central explanatory claim---that crashes and fat tails are endogenous, generic consequences of complex-systems dynamics (self-organized criticality, phase transitions) rather than responses to identifiable external shocks \citep{bak_self-organized_1987, sornette_why_2003}---is in tension with event-driven explanations of the same phenomena from the news and behavioral-finance literatures (Sections~\ref{sec:news} and~\ref{sec:bf}), which trace specific crashes to specific informational or psychological triggers (earnings surprises, sentiment cascades, liquidity withdrawal). Both camps can point to real evidence: SOC-style models reproduce power-law-distributed event sizes without any exogenous large shock being assumed, which is consistent with crashes being generic features of the dynamics; but documented crashes (the 2010 Flash Crash, COVID-19 crash) are also retrospectively attributable to identifiable triggers (algorithmic feedback loops, a specific pandemic shock) \citep{sec_flash_2010, kirilenko_flash_2017}. The econophysics literature has not resolved whether these triggers are necessary causes or merely the proximate occasion for a crash that the system's internal dynamics made inevitable regardless of which specific shock arrived---a distinction that is empirically difficult to test because a real market only ever experiences one realized history. This is directly consequential for what a successful persona-bot simulation would prove: if the framework reproduces realistic crash statistics only when fed a historically-calibrated news shock (Section~\ref{sec:framework}, External Information Channel), that is evidence for the event-driven view; if it reproduces them even under unshocked, stationary information conditions, that would be evidence for the SOC view; the framework as currently specified is not designed to distinguish between these two outcomes, and Section~\ref{sec:validation} does not currently include a no-shock control condition that would let it do so.

More broadly, markets exhibit complex dynamics characteristic of far-from-equilibrium systems. Power laws, cascades, and emergent phenomena arise naturally when heterogeneous agents interact adaptively. Persona-trained simulations that capture heterogeneity and local interaction should generate realistic complex dynamics, including rare events and fat tails.

This is the central empirical claim PTMC must be validated against: fat tails and rare extreme events should arise emergently, not be imposed; see Table~\ref{tab:design-mapping}.

\subsection{Game Theory, Mechanism Design, and Strategic Interaction}
\label{sec:game-theory}

Game theory and mechanism design provide formal tools for strategic interaction and institution design. Foundational concepts---von Neumann and Morgenstern's rational decision-making under uncertainty \cite{von_neumann_theory_1944}, Nash equilibrium \cite{nash_equilibrium_1950}, Harsanyi's incomplete-information games \cite{harsanyi_games_1967}, and Aumann's correlated equilibrium \cite{aumann_rationality_1997}---underpin Myerson's mechanism-design framework for inducing truth-telling \cite{myerson_mechanism_1991} and Wilson's market-clearing algorithms \cite{wilson_equilibrium_1971}. Green and Laffont show fundamental trade-offs between efficiency, incentive compatibility, and individual rationality under private information \cite{green_information_1986}, and Schelling's micromotives/macrobehavior framework shows how individual constraints generate unintended macro-level phenomena \cite{schelling_micromotives_1978}. Multi-agent-systems theory \cite{weiss_multiagent_1999, wooldridge_agent-based_2009} formalizes how equilibrium in such settings depends on institutional structure and information, not agents alone.

Classical game theory and the bounded-rationality literature disagree about what a ``correct'' model of strategic reasoning in markets should require. The Nash/correlated-equilibrium tradition \citep{nash_equilibrium_1950, aumann_rationality_1997} treats equilibrium computation---each agent best-responding to correct beliefs about every other agent's strategy---as the normative benchmark against which real behavior should be measured, and deviations from it are treated as failures of rationality to be explained away. Evans and Prokopenko's result \citep{evans_bounded_2023} and the broader bounded-rationality tradition instead treat shallow, finite-depth strategic reasoning as the more empirically accurate \emph{positive} model of how real traders behave, not a degraded approximation to full equilibrium reasoning that real agents fail to reach. These are different scientific claims with different implications for this paper: the equilibrium tradition would predict that as bots are trained on more data and become better at reasoning about other bots, the simulation should converge toward Nash-equilibrium-consistent prices, whereas the bounded-rationality tradition predicts no such convergence is expected or desirable, since the empirically accurate target is shallow reasoning, not deep reasoning approximated. The literature does not adjudicate which prediction is correct for real markets at the depth of reasoning relevant to a persona-bot population, and the choice has a direct design consequence: if bots are trained purely via behavioral cloning on real trader actions (Section~\ref{sec:framework}), they will inherit whatever depth of strategic reasoning is implicit in the training data, but nothing in the literature tells us what that depth actually is or whether it differs systematically across the demographic strata the framework intends to represent.

In sum, game theory and mechanism design establish how individual incentives and institutional rules determine outcomes. Markets are institutions where decentralized decisions (via order matching) aggregate information. Persona-trained bots should encode strategic reasoning: each bot understands others' incentives and responds accordingly.

Evans and Prokopenko's finding that bounded, finite-depth strategic reasoning (not behavioral cloning itself, but a hand-specified cognitive-hierarchy model) is sufficient for realistic crisis dynamics \citep{evans_bounded_2023} supports training $\pi_\phi$ via behavioral cloning on real trader actions without separately imposing unbounded rationality; see Table~\ref{tab:design-mapping}.

\subsection{Mathematical Foundations: Stochastic Processes, Information Theory, and Statistical Physics}
\label{sec:math}

The classical Monte Carlo metaphor of Section~\ref{sec:framework} is grounded in stochastic process theory: geometric Brownian motion and its generalizations define the random paths averaged over in option pricing \citep{black_scholes_1973, merton_rational_1973}. PTMC replaces this exogenously specified process with an endogenous one: price paths emerge from aggregate bot order flow, approaching a mean-field limit as $K$ grows. Carmona and Lacker's probabilistic formulation of mean-field games \citep{carmona_meanfield_2015} characterizes the limiting behavior of large populations optimizing against the population's aggregate action distribution, offering a principled framework for the stability and convergence of the aggregate price process.

Glasserman, Mamaysky, and Qin's entropy-based news-novelty measure (Section~\ref{sec:news}) is an information-theoretic quantity \citep{glasserman_new_2023}, suggesting a validation extension: comparing the information flowing from news into bot order-flow decisions against empirically estimated transfer-entropy benchmarks, as a model-free check against under- or over-reaction. Bardoscia et al.'s statistical-physics treatment of financial networks \citep{bardoscia_physics_2021} supplies a vocabulary---order parameters, critical exponents, percolation thresholds---for characterizing when a simulation transitions from orderly price discovery into crisis-like cascade. Fabiani et al.'s ML surrogate models for identifying bifurcations in agent-based models from simulation data \citep{fabiani_tipping_2024} offer a diagnostic tool for locating cascade-prone parameter regions without full ensemble runs at every parameter setting, reducing the cost of the sensitivity analysis in Section~\ref{sec:validation}.

Mean-field game theory and persona heterogeneity are in tension over what does the explanatory work. Mean-field methods take the limit of infinitely many statistically interchangeable agents reacting to aggregate population behavior---a limit in which any individual agent's demographic profile and idiosyncratic history washes out, which is exactly what PTMC's value proposition depends on preserving. If the emergent phenomena of interest (crashes, volatility clustering) survive in the mean-field limit, the specific heterogeneity PTMC learns may be analytically dispensable for aggregate statistics, even if it remains useful for agent-level validation (Section~\ref{sec:validation}). Conversely, if the phenomena of interest are inherently finite-population effects---a single large institutional bot triggering a cascade, for instance---mean-field theory does not apply to the regime PTMC targets. At what population size a PTMC simulation would transition from a finite-population to a mean-field regime is an open question.

This supplies the mean-field scaling analysis, an information-theoretic validation criterion, and bifurcation-detecting surrogates used in Sections~\ref{sec:framework}--\ref{sec:validation}; see Table~\ref{tab:design-mapping}.

\subsection{Machine Learning Methodology: Architectures, Training, and Validation}
\label{sec:ml}

Beyond RL and LLMs specifically, several further ML subfields contribute design elements to persona-bot training. Statistical-learning foundations \cite{hastie_statistical_2009} and cross-validation procedures \cite{arlot_survey_2010} underpin behavioral cloning as supervised learning from trader actions. Graph neural networks---GCNs \cite{kipf_semi-supervised_2017}, relational inductive biases \cite{battaglia_relational_2018}, neural message passing \cite{gilmer_neural_2017}, expressiveness results \cite{xu_powerful_2019}, and surveys thereof \cite{wu_comprehensive_2021}---and neural-symbolic integration \cite{sankar_graph_2020} are candidate architectures if bot interaction is later modeled as a graph rather than a flat LOB. Graph and sequence embedding techniques \cite{goyal_graph_2020} and transfer learning/domain adaptation \cite{taylor_transfer_2009, rusu_sim-to-real_2016} are relevant to generalizing trained policies across markets (Section~\ref{sec:conclusion}). Variational/probabilistic inference \cite{tran_variational_2015} is a candidate technique for representing policy uncertainty.

Before gradient-based deep learning, evolutionary algorithms---genetic algorithms \cite{holland_genetic_1992}, genetic programming \cite{koza_genetic_1992}, and their finance applications \cite{chen_genetics_2012, dempster_algorithmic_2002, brabazon_bioinspired_2008}---were the dominant approach to learning agent behaviors, and the Santa Fe ASM (Section~\ref{sec:abe}) used exactly this to evolve trading rules. This is not merely historical: Such et al.\ show gradient-free genetic search over deep-network weights remains competitive with Q-learning and policy gradients at million-parameter scale \citep{such_deep_2017}, so evolutionary methods remain a live alternative for training or architecture-searching $\pi_\phi$ if gradient-based training proves unstable---differing from the Santa Fe ASM's heterogeneity mechanism only in whether heterogeneity is fitness-selected or learned from data.

Privacy-preserving training is the area of clearest direct relevance: federated learning \cite{mcmahan_communication-efficient_2017} and differential privacy \cite{dwork_differential_2006} are the leading mitigations for training on real trader records without centralizing or exposing them, surveyed by Kairouz et al.\ \cite{fedorov_federated_2021}; membership-inference \cite{nasr_comprehensive_2019, shokri_membership_2017} and model-inversion \cite{fredrikson_model_2015} attacks document the corresponding risk if these mitigations are skipped.

Differential privacy and behavioral fidelity pull in opposite directions---the noise required for a meaningful privacy guarantee degrades exactly the fine-grained individual behavioral signal the framework depends on---and the literature documents no theoretical result resolving this trade-off for persona-level training \citep{fedorov_federated_2021, dwork_differential_2006}.

On balance, modern machine learning provides powerful tools: graph neural networks for structured data, federated learning for privacy, transfer learning for domain adaptation, probabilistic inference for uncertainty. Persona-bot training should leverage these techniques to ensure behavioral fidelity while protecting trader privacy.

Federated learning and differential privacy are a structural requirement given Training Data Sources' reliance on real trader records; see Table~\ref{tab:design-mapping}.

\subsection{Empirical Calibration and Validation Against Stylized Facts}
\label{sec:cal}

A core challenge in ABM is validation: how do we know if our model is accurately capturing reality? The field has developed several approaches.

\textbf{Stylized-fact matching:}

Cont (2001) documented empirical regularities (stylized facts) of asset returns robust across markets and time periods \cite{cont_stylized_2001}. An ABM is validated if it reproduces these facts:
\begin{itemize}
  \item \textbf{Absence of autocorrelation in returns:} Daily returns are approximately white noise (random walk).
  \item \textbf{Heavy tails:} The empirical distribution of returns has fatter tails than a normal distribution (excess kurtosis > 0).
  \item \textbf{Volatility clustering:} High-volatility periods tend to cluster; the autocorrelation of absolute returns decays slowly.
  \item \textbf{Leverage effect:} Volatility tends to increase after negative returns and decrease after positive returns (negative correlation).
  \item \textbf{Long memory in volatility:} Absolute returns exhibit slow power-law decay in autocorrelation (fractional integration).
\end{itemize}

Andersen, Bollerslev, Diebold, and Ebens provide precise intraday benchmarks for these facts \cite{andersen_stylized_2006}, and LeBaron, Arthur, and Palmer demonstrated the Santa Fe ASM could be calibrated to reproduce them \cite{lebaron_calibration_1999}. Statistical confirmation relies on standard goodness-of-fit and normality tests---Anderson-Darling \cite{anderson_anderson-darling_1952}, Kolmogorov \cite{kolmogorov_sulla_1933}, Lilliefors \cite{lilliefors_kolmogorov-smirnov_1967}, and Jarque-Bera \cite{jarque_efficient_1987}.

Parameter calibration and validation methodology are addressed by a cluster of closely related contributions: Farmer and Foley flag the complexity/identifiability trade-off in ABM calibration \cite{farmer_calibration_2012}; Fagiolo, Moneta, and Windrum give a critical guide to calibration and robustness practice \cite{fagiolo_validation_2007}; Windrum, Fagiolo, and Moneta detail systematic sensitivity analysis and identification of load-bearing assumptions \cite{windrum_empirical_2007}; Fagiolo and Roventini argue for pre-specifying validation targets before calibration \cite{fagiolo_computational_2019}; and Balci's framework distinguishes validation (matching observed data) from verification (code matches intended model) \cite{balci_validation_1998}. Approximate Bayesian Computation \cite{beaumont_approximate_2002} and engineering-derived simulation validation/verification frameworks \cite{balci_validation_1998, sterman_validation_2002} extend this toolkit to likelihood-intractable ABMs, and Lamperti, Roventini, and Sani's ML-surrogate approach---training a fast supervised model to approximate the ABM's parameter-to-output mapping---substantially reduces the computational cost of searching this calibration space \citep{lamperti_calibration_2018}, directly motivating this paper's later claim that ML surrogates can make PTMC's own population-hyperparameter calibration tractable.

Time-series econometrics supplies the comparison machinery between simulated and real price dynamics: ARIMA \cite{box_time_1970}, ARCH/GARCH for volatility clustering \cite{engle_arch_1982, bollerslev_garch_1986}, regime-switching models for bull/bear dynamics \cite{hamilton_regime-switching_1989}, VARs and impulse-response analysis \cite{stock_vector_2001, sims_macroeconomics_1980}, Bayesian VARs \cite{litterman_forecasting_1986}, state-space models \cite{durbin_time_1960}, structural identification \cite{pagan_structural_1996}, theory-guided specification \cite{gregory_empirical_1996}, return predictability \cite{timmermann_predictability_2018}, and GMM estimation \cite{hansen_large_1996}. Long-range dependence, characterized by the Hurst exponent ($H>0.5$ indicating persistence) \cite{hurst_long_1951}, has likewise been documented in financial returns.

The equifinality problem is the single sharpest objection to the entire validation methodology proposed in Section~\ref{sec:validation}, and it must be addressed explicitly rather than minimized. Fagiolo, Guerini, Lamperti, Moneta, and Roventini's methodological synthesis \citep{fagiolo_validation_2019} and Fabretti's calibration study \citep{fabretti_calibrating_2013} both document \emph{equifinality}: structurally very different generative mechanisms---rational-expectations models with carefully chosen noise processes, zero-intelligence trading institutions \citep{gode_allocative_1993}, heterogeneous-agent switching models \citep{lux_volatility_1999}, and (by direct extension) persona-trained neural policies---can all reproduce the same empirical stylized facts (fat tails, volatility clustering, long memory) documented by Cont \citep{cont_stylized_2001}. If many mechanistically unrelated models converge on the same handful of low-order statistical regularities, then successfully matching those regularities is, on its own, only weak evidence that any one model's internal mechanism is the one actually operating in real markets. This is directly consequential for the present framework's central empirical claim: demonstrating that a persona-bot simulation reproduces fat tails and volatility clustering (Section~\ref{sec:validation}, Level 1) would \emph{not}, by the equifinality argument, constitute evidence that the bots' learned behavioral mechanisms are realistic, only that the simulation belongs to the broad class of models---now empirically known to be large---capable of producing these statistics by many different routes. Fagiolo et al.\ argue the methodological corrective is to validate at a finer grain than aggregate stylized facts: matching microstructure-level and agent-level behavior (not just the resulting macro statistics) reduces, though does not eliminate, the equifinality problem, because the number of mechanisms capable of matching detailed micro-level patterns is smaller than the number capable of matching macro statistics alone. This is precisely the logic underlying this paper's Level 2 (microstructure) and Level 3 (agent-level) validation tiers in Section~\ref{sec:validation}, but it should be stated plainly that even satisfying all four validation levels proposed there would not constitute proof that persona-trained Monte Carlo's specific mechanism (behaviorally-cloned, demographically-conditioned neural policies) is correct, only that it has not yet been falsified by the available empirical tests---a weaker and more honest claim than ``the model is validated.''

In sum, ABMs should be validated against empirical stylized facts. Parameter calibration can be automated using optimization and Bayesian inference. Systematic sensitivity analysis and robustness checks are essential. Time-series econometrics provides tools for characterizing and comparing simulated vs. real price dynamics.

ML surrogate models can reduce the cost of calibrating PTMC's population-level hyperparameters against stylized facts; see Table~\ref{tab:design-mapping}.

\subsection{Design Implications Across Literatures}
\label{sec:design-mapping}

Each subsection above closed with a pointer to this table, which collects the literature-to-design mapping in one place rather than repeating it twelve times. The Tension/Open Disagreement blocks in each subsection remain the substantive record of what is unresolved; this table records only what each literature contributes to a specific design element of PTMC (Section~\ref{sec:framework}).

\begin{table}[htb]
\centering
\footnotesize
\begin{tabular}{p{3.0cm}p{3.6cm}p{6.5cm}}
\toprule
\textbf{Literature} & \textbf{PTMC design element} & \textbf{Specific contribution} \\
\midrule
Classical theory (\S\ref{sec:classical}) & Null hypothesis / $\theta$ features & GBM is the degenerate case ($\pi_\phi \to$ i.i.d.); CAPM beta, risk-return trade-off feed $\theta$ \\
ACE / ABM (\S\ref{sec:abe}) & Methodological ancestor & Heterogeneity learned from data rather than hand-specified or GA-evolved \\
Microstructure (\S\ref{sec:mm}) & Market mechanism, $s_t$ & CDA/LOB matching engine; order-flow/depth features in market state \\
Behavioral finance (\S\ref{sec:bf}) & Persona profile $\rho$, $\lambda$; $\pi_\phi$ internal architecture & Loss aversion, herding, momentum, anchoring as latent dimensions; separate gain/loss value heads and herding-sensitivity parameters motivated by neuroeconomic evidence \\
Deep RL (\S\ref{sec:rl}) & Training objective & Hybrid behavioral-cloning/IRL loss balances reward-seeking vs.\ realism \\
News/sentiment (\S\ref{sec:news}) & External information channel & FinBERT-style text embedding, novelty/entropy signal, macro/index features \\
Networks/systemic risk (\S\ref{sec:networks}) & Bot subpopulations & Multiple $(\theta,\rho)$ subpopulations enable endogenous cascade dynamics \\
Econophysics (\S\ref{sec:econ-physics}) & Validation target & Fat tails/extreme events should emerge, not be imposed (Level 1) \\
Game theory (\S\ref{sec:game-theory}) & $\pi_\phi$ rationality assumption & Bounded, behaviorally-cloned reasoning suffices; no unbounded rationality needed \\
Math.\ foundations (\S\ref{sec:math}) & Scaling, validation, diagnostics & Mean-field limits as $K\to\infty$; entropy-based information-flow check; bifurcation surrogates \\
ML methodology (\S\ref{sec:ml}) & Privacy pipeline & Federated learning / differential privacy required given real trader-record dependence \\
Calibration (\S\ref{sec:cal}) & Calibration cost & ML surrogates reduce cost of searching population-hyperparameter space \\
\bottomrule
\end{tabular}
\caption{Literature-to-design-element mapping for PTMC, consolidating the per-subsection ``Connection to the proposed framework'' notes above.}
\label{tab:design-mapping}
\end{table}

\section{Proposed Framework: Persona-Trained Monte Carlo (PTMC)}
\label{sec:framework}

We now formalize the proposed framework for market simulation using trained neural-network personas.

\textbf{Epistemic status of this section.} What follows is a literature-derived target architecture from Section~\ref{sec:related}, not an implemented or tested system: nothing here has been built, trained, or run, and any numerical examples are illustrative only. Read the diagrams, equations, and algorithms below as a falsifiable implementation roadmap, not validated results; Section~\ref{sec:implementation-status} states what would be required to test it.

\subsection{Conceptual Overview and Foundations}

Classical Monte Carlo valuation replaces deterministic computation with random sampling: to value a European call option, one simulates many paths of stock prices drawn from a specified stochastic process (e.g., geometric Brownian motion) and averages the discounted payoff across paths \cite{cox_option_1979}. The final estimate converges to the true value as sample size increases.

Persona-trained Monte Carlo extends this metaphor to market simulation. Instead of assuming a fixed population of traders (e.g., $N=1000$ rational agents, $M=500$ noise traders), we treat trader heterogeneity as an unobserved distribution. We sample virtual bots from a learned distribution of plausible trader personas, where each bot's decision-making reflects patterns observed in real trader data.

Formally, let $\mathcal{P}$ denote the unknown distribution of market participant personas. Each persona $p \sim \mathcal{P}$ can be characterized by a tuple:
\[
p = (\theta, \rho, \pi)
\]
where:
\begin{itemize}
  \item $\theta$ is a demographic vector (age, wealth, experience, risk tolerance).
  \item $\rho$ is a behavioral profile (encoded in learned latent representations).
  \item $\pi$ is a trading policy mapping market state to action (buy/sell/hold, order size, order price).
\end{itemize}

We approximate $\pi(a \mid s, p)$ (the probability of action $a$ given state $s$ and persona $p$) via a neural network trained on historical trading data conditioned on trader-specific variables ($\theta$, $\rho$). Let $\pi_\phi(a \mid s, \theta, \rho)$ denote the learned policy, parameterized by $\phi$. The objective is:
\[
\phi^* = \arg\min_\phi \mathbb{E}_{(s, \theta, \rho, a) \sim \mathcal{D}} \left[ \ell(\pi_\phi(a \mid s, \theta, \rho), a) \right]
\]
where $\mathcal{D}$ is a dataset of real trading decisions paired with market state, trader demographics, and behavioral covariates, and $\ell$ is a loss function (e.g., cross-entropy for discrete actions).

A simulation run instantiates $K$ bots (each a copy of $\pi_\phi$ with sampled parameters $(\theta^{(k)}, \rho^{(k)})$), initializes them with a market state (prices, order book, fundamentals), runs $T$ time steps, and records aggregate price dynamics. Different runs correspond to different random draws of bot personas, similar to different Monte Carlo paths in classical simulation.

\textbf{The PTMC estimator.} Let $F$ denote a target functional of a simulated market path: examples include maximum drawdown, a tail index of the return distribution, the stationary distribution of the bid-ask spread, or the probability of a crash exceeding a given magnitude within a horizon $T$. A single PTMC run produces one path $\text{path}_i$ by (i) drawing a population of $K$ persona tuples $\{(\theta^{(k)}, \rho^{(k)})\}_{k=1}^K \sim \mathcal{P}$, and (ii) simulating $T$ steps of the resulting CDA, where at each step every bot samples an action $a_t^{(k)} \sim \pi_\phi(\cdot \mid s_t, \theta^{(k)}, \rho^{(k)})$. Repeating this $N_{\text{runs}}$ times over independent persona-population draws yields the Monte Carlo estimator
\[
\hat{\mu}_N = \frac{1}{N_{\text{runs}}} \sum_{i=1}^{N_{\text{runs}}} F(\text{path}_i),
\]
with the usual Monte Carlo standard error $\hat{\sigma}/\sqrt{N_{\text{runs}}}$ and convergence $\hat{\mu}_N \to \mathbb{E}_{\mathcal{P}}[F]$ as $N_{\text{runs}} \to \infty$, holding $K$ fixed; a secondary convergence question, not addressed by classical Monte Carlo theory, is whether and how $\mathbb{E}_{\mathcal{P}}[F]$ itself varies as $K \to \infty$ (Section~\ref{sec:math} discusses the mean-field limit relevant to this second question).

PTMC's randomness therefore has three distinct sources, against exactly one in classical option-pricing Monte Carlo: (1) the outer persona-population draw $\{(\theta^{(k)}, \rho^{(k)})\}_{k=1}^K \sim \mathcal{P}$, which has no analogue in classical Monte Carlo; (2) within-run stochastic action sampling $a_t^{(k)} \sim \pi_\phi(\cdot \mid s_t, \theta^{(k)}, \rho^{(k)})$, analogous to the stochasticity of a single classical Monte Carlo path; and (3) an optional exogenous shock process (e.g., an injected news event or liquidity shock), which is the \emph{only} source of randomness in classical Monte Carlo applied to GBM. This is the precise sense in which PTMC differs from the four adjacent paradigms introduced in Section~\ref{sec:intro}: classical Monte Carlo has source (3) only; hand-coded ABMs have source (2) only, since the population is fixed rather than drawn from a learned $\mathcal{P}$; a single RL trading agent optimizes one policy and reports no ensemble at all; and LLM-based generative agents replace $\pi_\phi$ with language-model inference, retaining an analogue of sources (1)-(2) but losing the compact, shared, and reproducible structure that makes the estimator above cheap to evaluate at scale.

\begin{figure}[htb]
\centering
\resizebox{\textwidth}{!}{%
\begin{tikzpicture}[
  font=\small,
  block/.style={draw, rectangle, rounded corners, align=center, minimum width=5.6cm, minimum height=1.1cm, inner sep=4pt},
  src/.style={draw, rectangle, rounded corners, align=center, minimum width=2.6cm, minimum height=1.0cm, inner sep=3pt},
  arr/.style={->, >=latex, thick}
]
  % Row 1: four training-data sources (Sec. 3.5)
  \node[src] (d1) at (-7.5, 8.8) {Real transactions\\(Sec.~3.5)};
  \node[src] (d2) at (-2.5, 8.8) {Surveys \&\\experiments};
  \node[src] (d3) at (2.5, 8.8) {Synthetic\\(GAN/VAE)};
  \node[src] (d4) at (7.5, 8.8) {Transfer\\learning};

  % Row 2: feature engineering
  \node[block] (feat) at (0, 6.9) {Feature engineering and input\\normalization (Sec.~3.3)};
  \draw[arr] (d1.south) |- (feat.north);
  \draw[arr] (d2.south) |- (feat.north);
  \draw[arr] (d3.south) |- (feat.north);
  \draw[arr] (d4.south) |- (feat.north);

  % Row 3: training pipeline
  \node[block] (train) at (0, 5.0) {Policy training: behavioral cloning\\$\to$ inverse RL $\to$ hybrid fine-tuning\\(Sec.~3.7, Alg.~\ref{alg:bc}--\ref{alg:hybrid})};
  \draw[arr] (feat.south) -- (train.north);

  % Row 4: trained policy
  \node[block] (policy) at (0, 2.6) {Trained policy network $\pi_\phi^*$};
  \draw[arr] (train.south) -- (policy.north);

  % Row 5: persona sampling + simulation
  \node[src] (sample) at (-6.5, 0.4) {Draw persona\\population\\$\{(\theta^{(k)},\rho^{(k)})\}_{k=1}^K \sim \mathcal{P}$};
  \node[block, minimum width=4.6cm] (sim) at (0, 0.4) {Simulate continuous double\\auction, $T$ time steps (Sec.~3.6)};
  \node[src, minimum width=2.0cm] (path) at (6.5, 0.4) {path$_i$};
  \draw[arr] (policy.south) |- (sim.north);
  \draw[arr] (sample.east) -- (sim.west);
  \draw[arr] (sim.east) -- (path.west);

  % Row 6: Monte Carlo aggregation
  \node[block] (mc) at (0, -1.8) {Repeat over $N_{\text{runs}}$ independent persona\\draws: $\hat\mu_N = \frac{1}{N_{\text{runs}}}\sum_{i=1}^{N_{\text{runs}}} F(\text{path}_i)$};
  \draw[arr] (path.south) |- (mc.east);
  \draw[arr] (sample.south) |- (mc.west);

  % Row 7: validation
  \node[block] (val) at (0, -3.8) {Validation Levels 1--4 and PTMC-specific\\tests (Sec.~4)};
  \draw[arr] (mc.south) -- (val.north);

  % Exit branch: validation passes -> terminal output
  \node[block] (exit) at (0, -5.6) {Validated PTMC estimates $\hat\mu_N$\\(ready for use)};
  \draw[arr] (val.south) -- (exit.north) node[midway, right, font=\scriptsize] {pass};

  % Feedback loop: validation fails -> recalibrate training data / persona prior
  \draw[arr] (val.west) -- ++(-9.2,0) coordinate (loopx) |- (d1.west)
    node[pos=0.02, above, font=\scriptsize] {fail};
\end{tikzpicture}%
}
\caption{End-to-end PTMC pipeline. Training data from four candidate sources (Section~\ref{sec:framework}, Training Data Sources) are converted into normalized features (Section~\ref{sec:framework}, Feature Engineering) and used to train the shared policy network $\pi_\phi$ (Section~\ref{sec:framework}, Training Objectives). At simulation time, independent persona-population draws from $\mathcal{P}$ instantiate $K$ copies of $\pi_\phi^*$ inside a continuous double auction (Section~\ref{sec:cda}), producing one path per draw; repeating this over $N_{\text{runs}}$ draws yields the PTMC Monte Carlo estimator $\hat\mu_N$. Section~\ref{sec:validation}'s four-level validation protocol and PTMC-specific tests assess the resulting simulated paths: if validation passes, the pipeline exits with $\hat\mu_N$ ready for downstream use; if it fails, the persona prior or training data are recalibrated and the pipeline re-enters at its first stage.}
\label{fig:ptmc-workflow}
\end{figure}
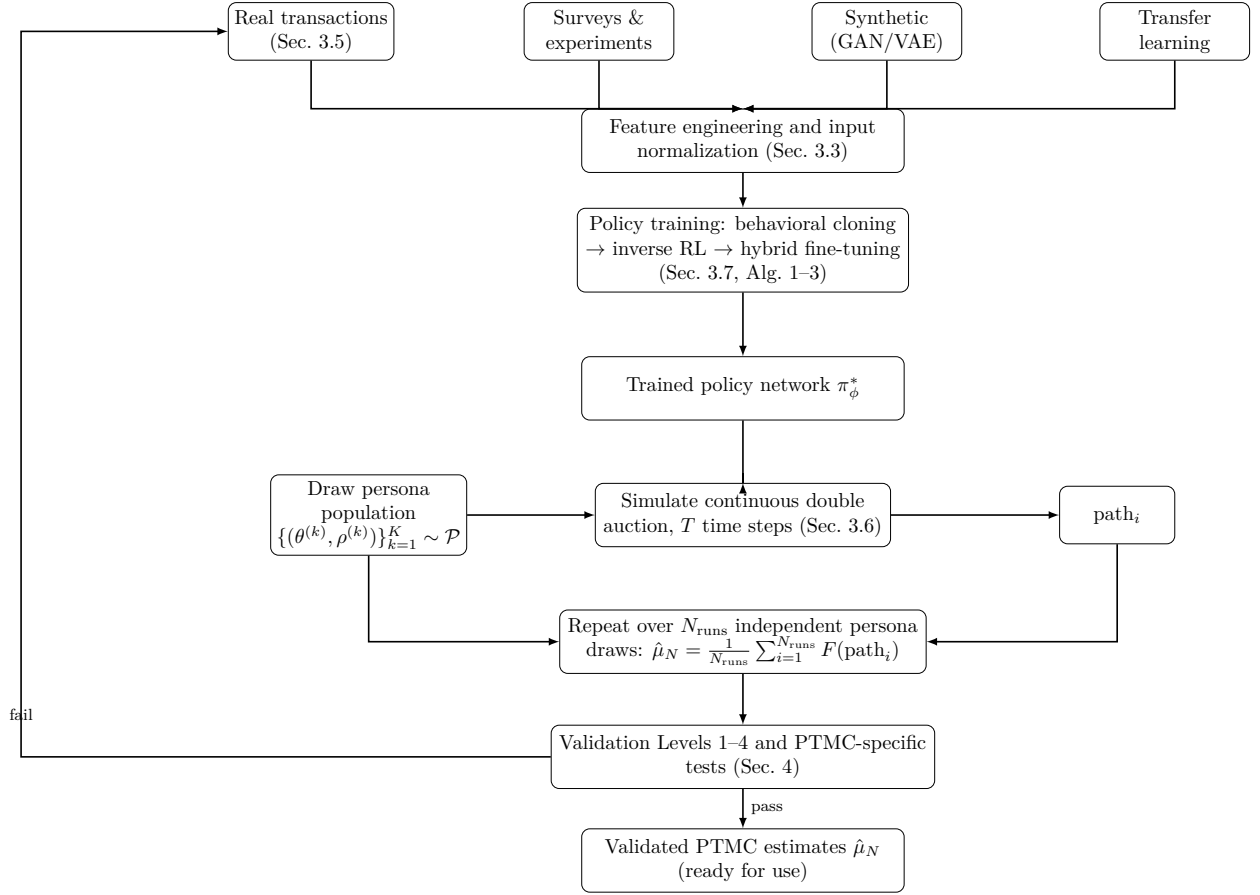

Figure~\ref{fig:ptmc-workflow} summarizes how the remaining subsections of this section and Section~\ref{sec:validation} fit together as a single end-to-end pipeline, from raw training data through to validation, with calibration feeding back into the training-data stage.

\subsection{Bot Agent Architecture}

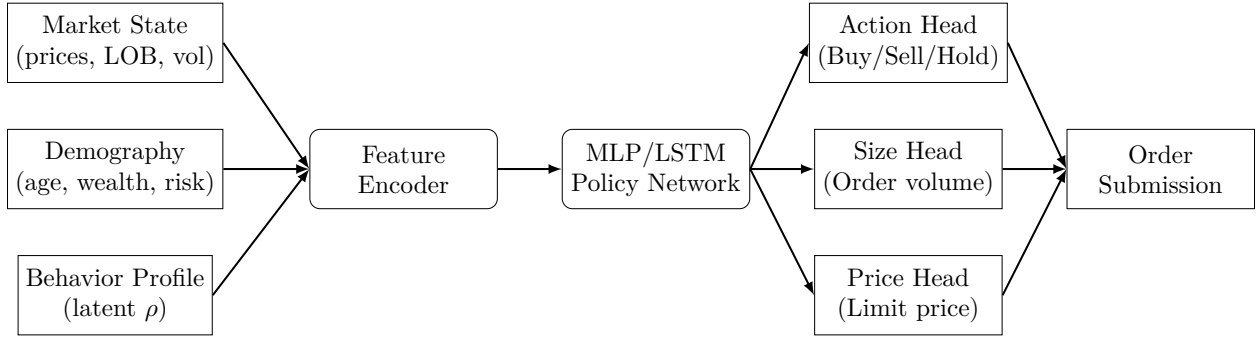
\begin{figure}[htb]
\centering
\resizebox{\textwidth}{!}{%
\begin{tikzpicture}[
  font=\small,
  block/.style={draw, rectangle, align=center, minimum width=2.6cm, minimum height=1.1cm, inner sep=3pt},
  enc/.style={block, rounded corners},
  arr/.style={->, >=latex, thick}
]
  % Input nodes
  \node[block] (market) at (0, 3.5) {Market State\\(prices, LOB, vol)};
  \node[block] (demo) at (0, 1.75) {Demography\\(age, wealth, risk)};
  \node[block] (behavior) at (0, 0) {Behavior Profile\\(latent $\rho$)};

  % Feature processing
  \node[enc] (feat) at (4, 1.75) {Feature\\Encoder};
  \draw[arr] (market.east) -- (feat.west);
  \draw[arr] (demo.east) -- (feat.west);
  \draw[arr] (behavior.east) -- (feat.west);

  % Core neural network
  \node[enc] (mlp) at (7.5, 1.75) {MLP/LSTM\\Policy Network};
  \draw[arr] (feat.east) -- (mlp.west);

  % Output heads
  \node[block] (action) at (11, 3.5) {Action Head\\(Buy/Sell/Hold)};
  \node[block] (size) at (11, 1.75) {Size Head\\(Order volume)};
  \node[block] (price) at (11, 0) {Price Head\\(Limit price)};

  \draw[arr] (mlp.east) -- (action.west);
  \draw[arr] (mlp.east) -- (size.west);
  \draw[arr] (mlp.east) -- (price.west);

  % Outputs
  \node[block] (order) at (14.5, 1.75) {Order\\Submission};
  \draw[arr] (action.east) -- (order.west);
  \draw[arr] (size.east) -- (order.west);
  \draw[arr] (price.east) -- (order.west);
\end{tikzpicture}%
}
\caption{Architecture of a persona-bot agent. Market state, demographics, and learned behavioral profiles feed into a neural-network policy, which outputs a trading action (order type, size, and price).}
\label{fig:bot-arch}
\end{figure}

Figure \ref{fig:bot-arch} illustrates the architecture of a single persona-bot.

A bot ingests three types of inputs. The market state $s_t$ comprises the observable market configuration at time $t$: the last trade price, the bid-ask midpoint and spread, limit-order-book depth (prices and volumes at each level, e.g., the top five bid/ask levels), realized volatility (e.g., a rolling 30-day standard deviation of returns), trading volume and order-flow imbalance, and, optionally, news sentiment and macro variables (the yield curve, VIX, Fed rates). Demographics and preferences $\theta^{(k)}$ are bot-specific parameters: age, the wealth or capital allocated to trading, experience or tenure in the market, a risk-aversion coefficient $\gamma$ (from a utility function or behavioral elicitation), a time horizon (short-term day trader vs.\ long-term investor), and a loss-aversion coefficient $\lambda$ from prospect theory \cite{kahneman_prospect_1979}. The behavioral profile $\rho^{(k)}$ consists of learned latent variables encoding the persona's decision-making style---herding tendency (the propensity to follow other traders), momentum bias (a tendency to buy uptrends and sell downtrends), anchoring bias (reliance on recent prices rather than fundamentals), and disposition bias (reluctance to realize losses)---and is either (a) learned via variational autoencoders (VAEs) \cite{kingma_auto-encoding_2014} or principal component analysis (PCA) from trader-behavior data, or (b) sampled from a prior distribution fitted to behavioral surveys and experiments.

The policy network $\pi_\phi$ processes these inputs through a feature encoder that projects market state, demographics, and behavior profile into a shared embedding; one or more hidden layers, potentially with recurrence (e.g., an LSTM), to capture temporal dependencies; and multiple output heads that produce an action distribution (logits over Buy, Sell, and Hold, or a more granular action set), an order size as a continuous distribution (e.g., log-normal) over order volume, and an order price as a relative offset from the current bid-ask midpoint (e.g., placing the order $\Delta p_t$ above or below the midpoint).

At each time step the bot observes $s_t$ and computes $(\pi_\phi, \sigma_\phi) = \text{policy\_net}(s_t, \theta, \rho)$, the mean and standard deviation of the action distribution; samples an action $a_t \sim \pi_\phi(\cdot \mid s_t, \theta, \rho)$ together with an order size $q_t$ and price $p_t$; submits the resulting order to the market mechanism (the continuous double-auction matching engine); and receives feedback in the form of order execution, realized profit and loss, and updated wealth.

\subsection{Feature Engineering and Input Normalization}

Effective feature engineering is critical for neural-network training and generalization. The raw market state ($s_t$) and demographics ($\theta$) must be converted into normalized numerical features suitable for supervised learning.

Raw limit-order-book data are high-dimensional and non-stationary, so several preprocessing steps are recommended for market-state features. Prices should be normalized to log-returns relative to a baseline (e.g., the first observed price), which makes the features scale-invariant and stationary; order-book depths should similarly be normalized by average daily volume, yielding dimensionless depth ratios. Rather than storing prices and volumes at all 50-plus LOB levels, the book should be compressed into summary statistics: cumulative volume within $K$ basis points of the midpoint (e.g., bid volume in $[-10\text{bp}, 0]$ and ask volume in $[0, +10\text{bp}]$), the spread (best ask minus best bid), and depth imbalance (bid volume divided by total volume). Temporal features---lagged returns, volatility (a rolling standard deviation), and autocorrelations---should also be computed; for example, $r_{t-1}, r_{t-2}, \ldots, r_{t-10}$ (the past 10 returns) and $\sigma_t$ (realized volatility over the past 30 periods) capture trend and volatility information. Finally, order-flow features such as buy-initiated versus sell-initiated trade volume (proxied by whether trades occurred above or below the midpoint), order imbalance (net buy volume divided by total volume), and momentum indicators (e.g., VWAP-adjusted returns) are strong predictors of future price moves and provide herding signals.

Demographic and behavioral features are often discrete or categorical and require their own encoding. Age should be normalized to a $[0,1]$ range (e.g., $(\text{age}-25)/(75-25)$), wealth log-scaled to handle its wide range, and experience normalized in years; risk-aversion $\gamma$ and loss-aversion $\lambda$ are typically elicited from surveys or experiments and are already numerical. Categorical variables such as time horizon (e.g., day trader = 0, swing trader = 1, long-term investor = 2) can be one-hot or, where a natural ordering exists, ordinally encoded, and professional-versus-retail status can be encoded as a binary indicator. The behavioral profile $\rho$ is harder to specify a priori: where sufficient trader-behavior time-series data is available, a low-dimensional representation can be learned via a VAE (Kingma and Welling, 2014), compressing a window of raw trading decisions (e.g., the past 100, $\{a_{t-100}, \ldots, a_{t-1}\}$) into a 2--5 dimensional latent vector $\rho$ that captures overall trading style without hand-specifying its dimensions; alternatively, PCA can be applied to a large matrix of trader-by-decision features.

After this feature engineering, the individual components are concatenated into a single input vector:
\[
\mathbf{x}_t = [\mathbf{f}_{\text{market}}(s_t); \mathbf{f}_{\text{demo}}(\theta); \mathbf{f}_{\text{behavior}}(\rho)]
\]
where $\mathbf{f}_{\text{market}} \in \mathbb{R}^{d_m}$, $\mathbf{f}_{\text{demo}} \in \mathbb{R}^{d_d}$, $\mathbf{f}_{\text{behavior}} \in \mathbb{R}^{d_b}$, and the concatenated vector $\mathbf{x}_t \in \mathbb{R}^{d_m + d_d + d_b}$ (typically 50--200 dimensions). Standardize this vector to zero mean and unit variance before passing to the neural network.

Following standard supervised-learning guidance \cite{hastie_statistical_2009}, training should start from a few strong features (last price, spread, order imbalance, risk-aversion) and add complexity only where cross-validation accuracy improves; avoid information leakage from future returns into features computed at time $t$; regularize (L1/L2, dropout) against overfitting to weak features; and validate that test-set feature distributions match training, since features rare in training (e.g.\ 50\% spreads) will generalize poorly to regimes where they are common.

\subsection{External Information Channel: Multimodal News, Macro, and Index Inputs}
\label{sec:external-info}

The architecture described above (Bot Agent Architecture, Feature Engineering) conditions each bot's decision on market state, demographics, and behavioral profile. This is insufficient to represent real market participants, who also consume and react to financial news, macroeconomic releases, sector-level trends, and index-level signals---a channel of information entirely distinct from price/order history. Section~\ref{sec:news} surveyed the empirical evidence that news positioning \citep{fedyk_frontpage_2024}, macro-micro news interaction \citep{hirshleifer_macro_2022}, and news novelty \citep{glasserman_new_2023} are first-order determinants of trading behavior; this subsection specifies how that evidence is operationalized as an additional input channel to $\pi_\phi$.

\textbf{Architecture of the external information channel:}

We extend the bot input vector $\mathbf{x}_t$ (Section~\ref{sec:framework}, Feature Engineering) with a fourth component, $\mathbf{f}_{\text{info}}(\mathbf{n}_t)$, derived from a structured external-information observation $\mathbf{n}_t$ at time $t$:

\[
\mathbf{x}_t = [\mathbf{f}_{\text{market}}(s_t); \mathbf{f}_{\text{demo}}(\theta); \mathbf{f}_{\text{behavior}}(\rho); \mathbf{f}_{\text{info}}(\mathbf{n}_t)]
\]

The observation $\mathbf{n}_t$ has three components:

\begin{itemize}
  \item \textbf{News text embedding:} Raw news headlines and article text relevant to the simulated security (and to the broader market) are encoded using a domain-adapted transformer language model in the FinBERT family \citep{araci_finbert_2019}, yielding a fixed-dimensional embedding vector $\mathbf{e}_{\text{news}} \in \mathbb{R}^{d_n}$ for each news item observed within the simulation step. Multiple simultaneous news items are aggregated via attention-weighted pooling, where attention weight is modulated by an estimated salience score informed by Fedyk's finding that news prominence (not content alone) drives reaction \citep{fedyk_frontpage_2024}.
  \item \textbf{News novelty signal:} In parallel with the content embedding, we compute a scalar (or low-dimensional) novelty score $\nu_t$ using the entropy-based methodology of Glasserman, Mamaysky, and Qin \citep{glasserman_new_2023}: the divergence between the distribution of language in current news and a rolling window of recent news. This is included as a separate feature because novelty carries predictive information independent of sentiment polarity, per their empirical finding that entropy outperforms standard sentiment measures as an out-of-sample predictor of returns.
  \item \textbf{Macro/sector/index trend vector:} A numerical vector $\mathbf{m}_t$ summarizing macro-level state: relevant index levels and recent returns, sector-level momentum, implied volatility (e.g., VIX-style measures), yield-curve features, and other macro indicators already noted as optional market-state features in Feature Engineering. Distinguishing this from firm-specific news operationalizes the macro/micro distinction documented by Hirshleifer and Sheng \citep{hirshleifer_macro_2022}: bots receive macro and micro information as separate, separately-weighted streams rather than a single blended signal.
\end{itemize}

The combined external-information feature is $\mathbf{f}_{\text{info}}(\mathbf{n}_t) = [\mathbf{e}_{\text{news}}; \nu_t; \mathbf{m}_t]$, concatenated into the overall input vector exactly as market-state, demographic, and behavioral features are. The feature encoder (Section~\ref{sec:framework}, Bot Agent Architecture) is correspondingly extended with an additional encoder branch for $\mathbf{f}_{\text{info}}$, allowing the shared embedding layer to learn cross-modal interactions between, for instance, a bot's herding tendency ($\rho$) and its sensitivity to high-novelty news ($\nu_t$).

\textbf{Training implications:}

Training data for the news-embedding branch requires paired (news, subsequent trader action) examples, which can be constructed by time-aligning historical news feeds with the same trader-level transaction data used elsewhere (Training Data Sources, below). Where such paired data is unavailable, the FinBERT encoder can be pretrained on general financial-sentiment corpora \citep{araci_finbert_2019} and fine-tuned end-to-end with the rest of $\pi_\phi$ using the hybrid behavioral-cloning/inverse-RL objective of Section~\ref{sec:framework}, allowing the network to learn the magnitude of news-driven behavioral adjustment directly from data rather than assuming a fixed sentiment-to-action mapping.

\textbf{Validation implications:}

Section~\ref{sec:math} proposed an information-theoretic validation criterion: comparing the rate at which information from $\mathbf{f}_{\text{info}}$ propagates into simulated order-flow decisions against empirically estimated news-to-price information-transfer rates in real markets. This is added as an explicit additional check within the Level 1 stylized-fact validation protocol of Section~\ref{sec:validation}, alongside the existing return-distribution and volatility-clustering checks, and is essential for confirming that bots are using external information at a realistic rate rather than ignoring it or overreacting to every incoming signal.

\subsection{Training Data Sources and Behavioral Calibration}

To train $\pi_\phi$, a dataset of tuples $(s_t, \theta_i, \rho_i, a_t)$ is needed, where $a_t$ is the action taken by trader $i$ at time $t$ given state $s_t$. Four candidate sources of such data exist. The first is real transaction data with trader identifiers: regulatory filings (SEC Form 4 for insider trades, large-position reports) and broker-provided datasets include trader demographics alongside order and execution records, though privacy concerns limit availability and academic partnerships with exchanges or brokers would be required to obtain anonymized versions. The second is behavioral experiments and surveys: laboratory experiments (e.g., Gode and Sunder's continuous-double-auction experiments, Smith's double-auction studies) provide controlled data on trader behavior under varying conditions, and surveys of retail and professional traders can elicit risk preferences, loss aversion \cite{kahneman_prospect_1979, tversky_loss_1991}, herding tendency, and overconfidence via quantitative questionnaires such as Kahneman-Tversky gambling scenarios. The third is simulated data with imposed realism: a generative model (a GAN or VAE) trained on observed stylized facts of price dynamics and trader demographics can generate synthetic trading decisions, allowing scaling without privacy concerns, though this requires careful validation that the generated behavior is realistic; Goodfellow et al.\ (2014) and the subsequent GAN literature provide the relevant foundations \cite{goodfellow_generative_2014}. The fourth is transfer learning from related domains: behavioral data from e-commerce (herding in online reviews), social media (information cascades), or voting contexts could in principle be transferred to financial trading, building on Taylor and Stone's (2009) review of transfer learning and Rusu et al.'s (2017) work on sim-to-real transfer \cite{taylor_transfer_2009, rusu_sim-to-real_2016}.

A practical training corpus might combine all four sources: real transaction data for high-confidence base patterns, experimental data for behavioral calibration, synthetic data for augmentation, and transfer learning for domains lacking direct finance data. Federated learning approaches (McMahan et al., 2017; Kairouz et al., 2021) can train models across institutions while preserving privacy \cite{mcmahan_communication-efficient_2017, fedorov_federated_2021}.

Table~\ref{tab:data-sources} compares the four training-data options on the dimensions most relevant to a practical implementation decision.

\begin{table}[htb]
\centering
\small
\begin{tabular}{p{3.0cm}p{2.6cm}p{3.0cm}p{2.5cm}p{2.7cm}}
\toprule
\textbf{Data Source} & \textbf{Privacy Risk} & \textbf{Real-World Availability Today} & \textbf{Granularity} & \textbf{Realism} \\
\midrule
Real transaction data with trader ID & High---directly re-identifiable; subject to GDPR/CCPA and brokerage confidentiality & Low---requires regulatory or brokerage partnership, IRB approval, anonymization pipeline & Very high (individual order/execution level) & Very high (ground truth) \\
\addlinespace
Behavioral experiments and surveys & Low-Medium---typically consented, aggregated, but still individually identifiable in raw form & Medium---academic CDA experiments and survey panels exist (Gode-Sunder, Smith-style replications) but are small-sample & Medium (controlled conditions, not live markets) & Medium (lab behavior may not transfer to live high-stakes trading) \\
\addlinespace
Synthetic/generative data (GAN/VAE) & Very low---no real individuals represented, if generator is well-calibrated & High---can be generated on demand once a generator is trained & Adjustable, but only as granular as the generator's training data allowed & Uncertain---realism is bounded by and cannot exceed the realism of whatever real data trained the generator \\
\addlinespace
Transfer learning from related domains (e-commerce, social media, voting) & Low-Medium---depends on source domain's own privacy posture & High---large behavioral datasets already exist in these domains & High within source domain, uncertain after transfer to finance & Low-Medium---domain shift from e.g.\ review-herding to order-driven trading is unvalidated \\
\bottomrule
\end{tabular}
\caption{Comparison of candidate training-data sources for persona-bot calibration. No combination of these sources has been assembled or tested for this framework; the comparison is intended to clarify the practical trade-offs an implementation would face, particularly the privacy-realism trade-off already raised as an open tension in Section~\ref{sec:ml}.}
\label{tab:data-sources}
\end{table}

\subsection{Market Mechanism: Continuous Double Auction with Limit Order Book}
\label{sec:cda}

To generate aggregate price dynamics, we adopt a realistic market-clearing mechanism: the continuous double auction (CDA) with limit-order-book (LOB) matching:

\begin{itemize}
  \item Each bot submits limit orders: $(\text{side}, \text{price}, \text{volume})$, e.g., ``buy 100 shares at \$50.00''.
  \item Orders rest on the LOB until matched or cancelled.
  \item When a new order arrives, the matching engine executes it against standing orders in price-time priority (highest bid, lowest ask; ties broken by arrival time).
  \item A transaction occurs when bid $\geq$ ask; the execution price is typically the standing order's limit price (not the incoming order's price).
  \item After each match, the bid-ask spread and depths update; this new state feeds into the next round of bot decisions.
\end{itemize}

The LOB state at time $t$ is:
\[
\text{LOB}_t = \{(\text{bid}_1, v_1), (\text{bid}_2, v_2), \ldots, (\text{ask}_1, v_1), (\text{ask}_2, v_2), \ldots\}
\]
where bid prices are ordered descending and ask prices are ordered ascending.

At each discrete time step (e.g., 1 minute):
\begin{enumerate}
  \item Each bot decides on an order conditional on the current LOB state.
  \item Orders are collected (both new orders and cancellations from prior steps).
  \item The matching engine processes orders in a specified sequence (e.g., best bid first, then best ask, etc.).
  \item Matched orders are removed; unmatched orders remain on the LOB.
  \item Prices and depths update.
  \item Bots observe the new state and prepare for the next time step.
\end{enumerate}

This mechanism is faithful to modern electronic markets (NASDAQ, NYSE, EUREX) and has been extensively studied in the microstructure literature (Parlour and Seppi, 2008; Hasbrouck, 2007; Harris, 2003) \cite{parlour_limit_2008, hasbrouck_asset_2007, harris_microstructure_2003}.

\subsection{Training Objectives and Algorithms: Behavioral Cloning, Inverse RL, and Hybrid Loss}

Training $\pi_\phi$ requires a loss function; four objectives can be combined. \textbf{Behavioral cloning (BC)}, $\mathcal{L}_\text{BC} = -\mathbb{E}_{(s,\theta,\rho,a)\sim\mathcal{D}}[\log \pi_\phi(a\mid s,\theta,\rho)]$, directly matches observed trader actions via cross-entropy, but is prone to compounding distribution-mismatch errors when training data is limited. \textbf{Reward-based RL}, $\mathcal{L}_\text{RL} = -\mathbb{E}[\sum_t \gamma^t r_t \mid \pi_\phi]$ \cite{sutton_policy_2000}, encourages adaptive, profitable behavior but requires simulating a full market environment during training. \textbf{Inverse RL} infers the unknown reward function that rationalizes observed actions, bridging the two. \textbf{Persona-fidelity regularization}, $\mathcal{L}_\text{reg} = \|(\theta,\rho) - \mathbb{E}_{\mathcal{D}}[(\theta,\rho)]\|^2$, keeps learned personas close to the observed demographic/behavioral distribution, preventing mode collapse. The joint objective $\mathcal{L}_\text{total} = \alpha\mathcal{L}_\text{BC} + \beta\mathcal{L}_\text{RL} + \gamma\mathcal{L}_\text{reg}$ would typically start at $\beta=0$ (pure cloning) before fine-tuning with $\beta>0$; as Section~\ref{sec:rl}'s Tension block notes, no result in the cited literature establishes what weighting jointly satisfies behavioral fidelity and profit-seeking adaptation.

Algorithm~\ref{alg:bc} below implements behavioral cloning; Algorithm~\ref{alg:irl} extends it with inverse RL, inferring traders' implicit objective rather than copying actions directly, with an imitation-consistency constraint (Step 3) preventing the policy from diverging into unrealistic behavior during the RL phase; Algorithm~\ref{alg:hybrid} combines both into a four-phase pipeline (BC initialization, IRL reward inference, joint-loss fine-tuning, robustness checks).

\begin{algorithm}[htb]
\caption{Behavioral Cloning for Persona-Bot Training (candidate procedure, untested)}
\label{alg:bc}
\begin{algorithmic}[1]
\Require Dataset $\mathcal{D} = \{(s_t^{(i)}, \theta_i, \rho_i, a_t^{(i)}) : i \in [1, N_{\text{traders}}], t \in [1, T_i]\}$ of observed trading decisions
\Require Neural network architecture $\pi_\phi$ (MLP with feature encoder, hidden layers, output heads)
\Require Learning rate $\eta$, batch size $B$, epochs $E_{\text{max}}$, validation set $\mathcal{D}_{\text{val}}$
\Ensure Trained policy $\pi_\phi^*$ that reproduces observed trader actions

\State Initialize network parameters $\phi$ randomly
\State Initialize best validation loss $\text{loss}_{\text{best}} \gets \infty$
\For{epoch $e = 1$ to $E_{\text{max}}$}
  \State Shuffle $\mathcal{D}$ into minibatches of size $B$
  \For{each minibatch $\mathcal{B} \subset \mathcal{D}$}
    \State Forward pass: compute $\pi_\phi(a \mid s, \theta, \rho)$ for all $(s, \theta, \rho, a) \in \mathcal{B}$
    \State Compute cross-entropy loss: $\mathcal{L}_\text{BC} = -\frac{1}{|\mathcal{B}|} \sum_{(s, \theta, \rho, a) \in \mathcal{B}} \log \pi_\phi(a \mid s, \theta, \rho)$
    \State Backward pass: compute gradients $\nabla_\phi \mathcal{L}_\text{BC}$
    \State Update: $\phi \gets \phi - \eta \nabla_\phi \mathcal{L}_\text{BC}$
  \EndFor
  \State Evaluate on validation set: $\text{loss}_{\text{val}} = \frac{1}{|\mathcal{D}_{\text{val}}|} \sum_{(s, \theta, \rho, a) \in \mathcal{D}_{\text{val}}} -\log \pi_\phi(a \mid s, \theta, \rho)$
  \If{$\text{loss}_{\text{val}} < \text{loss}_{\text{best}}$}
    \State Store checkpoint: $\phi^* \gets \phi$, $\text{loss}_{\text{best}} \gets \text{loss}_{\text{val}}$
  \EndIf
  \State If no improvement for $N_{\text{patience}}$ epochs, exit early
\EndFor
\Return $\phi^*$
\end{algorithmic}
\end{algorithm}

Early stopping on validation loss prevents overfitting; top-1 held-out action accuracy is a practical performance metric, though any specific target (e.g., ``$>65\%$ for ternary action prediction'') is illustrative pending calibration against an actual trader-action dataset, not a benchmark established in the cited literature.

\begin{algorithm}[htb]
\caption{Inverse Reinforcement Learning for Persona-Bot Training (candidate procedure, untested)}
\label{alg:irl}
\begin{algorithmic}[1]
\Require Dataset $\mathcal{D}$ of observed trader actions and realized rewards (P\&L)
\Require Expert policy $\pi_E$ (from behavioral cloning)
\Require Reward function approximator $R_\psi$ (neural network)
\Require Policy learning algorithm (e.g., PPO)
\Require Learning rates $\eta_R$, $\eta_\pi$, max iterations $M$
\Ensure Inferred reward function $R_\psi^*$ and policy $\pi_\phi^*$ that explains observed behavior

\State Initialize reward function $\psi$ and policy $\phi$ (from behavioral cloning pre-training)
\For{iteration $m = 1$ to $M$}
  \State \textbf{Step 1 (Reward Inference):}
  \State Sample batch from $\mathcal{D}$: obtain trajectories $\tau^{(i)} = (s_1, a_1, s_2, a_2, \ldots, s_T, a_T)$ with realized rewards
  \State Compute reward predictions: $\hat{r}_t = R_\psi(s_t, a_t, \theta_i, \rho_i)$
  \State Minimize prediction error: $\mathcal{L}_R = \frac{1}{|\mathcal{B}|} \sum_{\tau \in \mathcal{B}} \| \hat{r}_t - r_t^{\text{realized}} \|^2$
  \State Update: $\psi \gets \psi - \eta_R \nabla_\psi \mathcal{L}_R$

  \State \textbf{Step 2 (Policy Optimization):}
  \State Use the inferred reward $R_\psi$ to train policy via policy gradient (PPO, A3C, or similar)
  \State Simulate $N_{\text{sim}}$ trajectories using current policy $\pi_\phi$ in a market environment
  \State Compute cumulative discounted rewards: $G_t = \sum_{t'=t}^T \gamma^{t'-t} R_\psi(s_{t'}, a_{t'})$
  \State Compute policy gradient and update: $\phi \gets \phi - \eta_\pi \nabla_\phi \mathcal{L}_\pi$

  \State \textbf{Step 3 (Constraint: Imitation Consistency):}
  \State Reweight with behavioral-cloning loss: ensure $\pi_\phi$ does not diverge too far from observed actions
  \State $\mathcal{L}_\text{combined} = (1-\lambda) \mathcal{L}_\pi + \lambda \mathcal{L}_\text{BC}$ where $\lambda \in [0.1, 0.3]$
\EndFor
\Return $\psi^*$, $\phi^*$
\end{algorithmic}
\end{algorithm}

\begin{algorithm}[htb]
\caption{Hybrid Training Pipeline: Behavioral Cloning + Inverse RL + Reward Augmentation (candidate procedure, untested)}
\label{alg:hybrid}
\begin{algorithmic}[1]
\Require Dataset $\mathcal{D}$, bot architecture $\pi_\phi$, hyperparameters $(\alpha, \beta, \gamma)$
\Ensure Trained policy $\pi_\phi^*$ combining behavioral realism with profit-seeking
\State \textbf{Phase 1: Behavioral Cloning (Initialize).}
\State Run Algorithm~\ref{alg:bc} to train $\pi_\phi$ on $\mathcal{D}$ until convergence
\State Store checkpoint: $\phi_{\text{BC}} \gets \phi^*$

\State \textbf{Phase 2: Inverse RL (Learn Reward Function).}
\State Initialize $R_\psi$ (reward predictor) with linear regression on $\mathcal{D}$: $R_\psi(s, a) \approx \text{realized\_PnL}(s, a)$
\State Run Algorithm~\ref{alg:irl} for $M = 50$--100 iterations
\State Store checkpoint: $\phi_{\text{IRL}} \gets \phi^*$

\State \textbf{Phase 3: Joint Loss Optimization (Fine-Tune).}
\State Resume training from $\phi_{\text{IRL}}$, but now use joint loss:
\[
\mathcal{L}_\text{total} = \alpha \mathcal{L}_\text{BC} + \beta \mathcal{L}_\text{RL} + \gamma \mathcal{L}_\text{reg}
\]
where $\alpha = 0.4$, $\beta = 0.5$, $\gamma = 0.1$ (weight toward RL, constrained by behavioral fidelity)
\State Run for $E_{\text{max}} = 20$--50 additional epochs with lower learning rate $\eta' = 0.1 \eta$
\State Validate on hold-out set: track both BC accuracy (action prediction) and RL performance (Sharpe ratio, max drawdown)

\State \textbf{Phase 4: Robustness Checks.}
\State Test on out-of-distribution market conditions (rare volatility regimes, stress periods)
\State Verify bot behavior is interpretable: high-loss-aversion bots should avoid losses, herding bots should respond to order-book imbalance
\Return Final policy $\phi^*$
\end{algorithmic}
\end{algorithm}

This staged pipeline is a candidate procedure, not a demonstrated result; nothing here has been run.

\subsection{Multi-Bot Simulation: A Stylized Market Session}

\textit{Illustrative only: invented bot parameters and scripted actions, not output from a running simulator.}

To demonstrate how individual bot decisions aggregate into market dynamics, consider a simplified scenario with $K=5$ agents, each conditioned on a distinct persona draw $(\theta^{(k)}, \rho^{(k)})$ as in Table~\ref{tab:illustrative-personas}.

\begin{table}[htb]
\centering
\small
\begin{tabular}{@{}rcccccc@{}}
\toprule
$k$ & \multicolumn{2}{c}{$\theta^{(k)}$} & \multicolumn{3}{c}{$\rho^{(k)}$ components} & \\
\cmidrule(lr){2-3}\cmidrule(lr){4-6}
 & Age & Wealth (\$) & $h^{(k)}$ & $m^{(k)}$ & $\lambda^{(k)}$ & ID \\
\midrule
1 & 35 & 100{,}000 & 0.9 & 0.7 & 1.5 & A \\
2 & 55 & 200{,}000 & 0.2 & 0.1 & 3.0 & B \\
3 & 30 & 50{,}000 & 0.5 & 0.85 & 1.0 & C \\
4 & 50 & 300{,}000 & 0.1 & 0.0 & 2.0 & D \\
5 & 28 & 20{,}000 & 0.95 & 0.8 & 2.5 & E \\
\bottomrule
\end{tabular}
\caption{Illustrative persona draws for $K=5$. Demographics $\theta^{(k)}=(\text{age}, \text{wealth})$; $(h^{(k)}, m^{(k)}, \lambda^{(k)})$ are herding, momentum bias, and loss-aversion scalars summarizing components of $\rho^{(k)}$. Column ID gives the agent label used in Table~\ref{tab:market-session}.}
\label{tab:illustrative-personas}
\end{table}

\begin{figure}[htb]
\centering
\begin{tikzpicture}
  \begin{axis}[
    width=6cm, height=4cm,
    xlabel={Time (minutes)},
    ylabel={Midpoint Price (\$)},
    grid=major,
    grid style={gray!30},
    xtick={0,1,2,3,4},
    ytick={50.00, 50.10, 50.20, 50.30},
    yticklabel style={/pgf/number format/.cd, fixed, precision=2},
    legend pos=outer north east
  ]
  \addplot[color=blue, thick, mark=o] coordinates {
    (0, 50.025)
    (1, 50.120)
    (2, 50.220)
    (3, 50.280)
    (4, 50.320)
  };
  \addlegendentry{Midpoint}

  \addplot[color=red, dashed, thin] coordinates {
    (0, 50.000)
    (4, 50.400)
  };
  \addlegendentry{Fundamental trend}
  \end{axis}
\end{tikzpicture}
\caption{Price path during the multi-bot market session. The solid blue line traces realized midpoint prices as bots submit orders. The dashed red line shows the theoretical fundamental value path (news-driven). The gap between realized and fundamental reflects overreaction due to herding, a behavioral anomaly. Volatility is highest at minute 1-2 (steepest slope) as ordering activity peaks.}
\label{fig:market-price-path}
\end{figure}
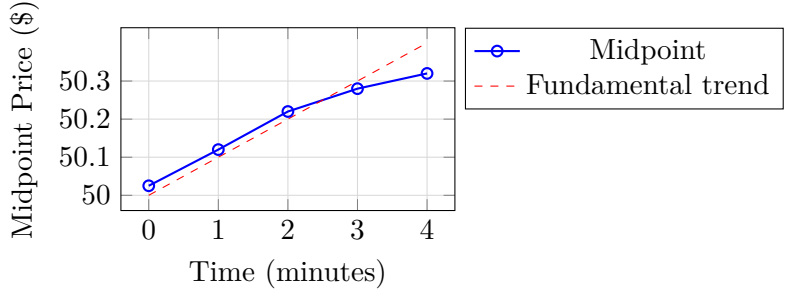

Initially, all agents hold equal dollar positions in an asset trading at \$50. The starting LOB shows balanced depth: 200 shares bid at \$50.00 and 200 shares ask at \$50.05. At $t=0$, a moderate positive news shock (e.g., an earnings beat) raises fundamental-value estimates by 2\%. Table~\ref{tab:market-session} and Figure~\ref{fig:market-price-path} trace the subsequent 5-minute session and resulting price dynamics.

\begin{table}[htb]
\centering
\small
\begin{tabular}{cccccccc}
\toprule
\textbf{Min} & \textbf{Bot} & \textbf{Action} & \textbf{Size} & \textbf{Price} & \textbf{Exec?} & \textbf{Mid Price} & \textbf{Vol.} \\
\midrule
0 & A & Buy & 50 & 50.05 & Partial (30) & \$50.025 & Baseline \\
  & B & Hold & $-$ & $-$ & No & & \\
  & C & Buy & 40 & 50.10 & Partial (20) & & \\
  & D & Hold & \textemdash & \textemdash & No & & \\
  & E & Buy & 35 & 50.08 & Partial (25) & & \\
\midrule
1 & A & Buy & 60 & 50.08 & Full & \$50.12 & 0.2\% \\
  & B & Hold & \textemdash & \textemdash & No & & rising \\
  & C & Buy & 50 & 50.15 & No (added to LOB) & & \\
  & D & Sell & 20 & 50.15 & No (above ask) & & \\
  & E & Buy & 40 & 50.10 & Partial (15) & & \\
\midrule
2 & A & Buy & 70 & 50.12 & Full & \$50.22 & 0.8\% \\
  & B & Sell & 15 & 50.20 & Partial (10) & & clust. \\
  & C & Buy & 55 & 50.18 & No & & \\
  & D & Hold & \textemdash & \textemdash & No & & \\
  & E & Buy & 50 & 50.15 & Full & & \\
\midrule
3 & A & Hold & \textemdash & \textemdash & No & \$50.28 & 0.3\% \\
  & B & Sell & 30 & 50.25 & Partial (20) & & momen. \\
  & C & Buy & 65 & 50.25 & Full & & \\
  & D & Sell & 40 & 50.18 & Full & & \\
  & E & Buy & 60 & 50.22 & Partial (40) & & \\
\midrule
4 & A & Buy & 80 & 50.30 & Partial (40) & \$50.32 & 0.2\% \\
  & B & Hold & \textemdash & \textemdash & No & & vol. \\
  & C & Sell & 30 & 50.35 & No & & cluster \\
  & D & Sell & 25 & 50.25 & Partial (10) & & \\
  & E & Buy & 45 & 50.28 & Partial (30) & & \\
\bottomrule
\end{tabular}
\caption{Illustrative multi-bot market session. A positive news shock triggers heterogeneous responses: momentum-biased bots (A, C, E) aggressively buy, while conservative/value bots (B, D) hold or sell. Order-book dynamics cause prices to rise from \$50.025 to \$50.32 in 5 minutes. Bot B's rational selling at peak (minute 2-3) is overwhelmed by herd buying. Volatility clustering emerges from correlated bot actions. This illustrates stylized facts (momentum, volatility clustering, herding) arising from heterogeneous learned policies.}
\label{tab:market-session}
\end{table}

\textbf{Key observations:} momentum/herding bots (A, C, E) buy immediately into the shock while conservative bots (B, D) hold or sell; the matching engine accumulates these orders into a rising execution price (\$50.05 $\to$ \$50.32), realizing the Kyle (1985) order-flow-aggregation mechanism \cite{kyle_continuous_1985}; volatility peaks at minutes 1-2 when buying is most intense and re-rises at minute 4 as bot C sells into the momentum, the temporal clustering structure central to stylized facts \cite{cont_stylized_2001} but absent from random-walk models; the aggregate price impact (0.6\% from five modestly-sized orders) reflects realistic nonlinear microstructure; and bot D's partially-filled sell at minute 3 suggests herding bots E and C's amplification of A's signal has limits, foreshadowing the mean-reversion anomaly an extended session would need to reproduce.

Figure~\ref{fig:market-price-path} visualizes the resulting price path: a positive trend with high-frequency micro-noise, contrasting with geometric Brownian motion's smooth paths.

This worked example demonstrates the core mechanism: heterogeneous, learned behavioral patterns drive simultaneous responses to market state, which interact through the matching engine to produce aggregate stylized facts absent from classical models. Replicating such sessions at scale forms the basis for validating the simulation against empirical markets, conditional on the implementation work described next.

\subsection{Implementation Status and What Would Be Required to Test This}
\label{sec:implementation-status}

\textbf{Data.} A working implementation requires, at minimum: (1) trader-level transaction records paired with demographic and behavioral covariates at a granularity matching Training Data Sources Option 1 (Section~\ref{sec:framework})---in practice this means a data-sharing agreement with a broker, exchange, or regulator, subject to IRB approval and the privacy-preserving training pipeline discussed in Section~\ref{sec:ml}; (2) a time-aligned historical news corpus covering the same trading period, for the external information channel (Section~\ref{sec:external-info}); and (3) historical limit-order-book data at the venue(s) being modeled, for microstructure validation (Section~\ref{sec:validation}, Level 2). None of these three data sources currently exists in assembled, linked form to our knowledge; assembling them is itself a multi-institution research undertaking independent of any modeling contribution.

\textbf{Compute.} Training $\pi_\phi$ via the hybrid behavioral-cloning/inverse-RL objective (Section~\ref{sec:framework}) at a scale sufficient to support thousands of simultaneously simulated, heterogeneous bots requires: GPU-accelerated training infrastructure comparable to standard deep-RL research setups; a vectorized, GPU- or multi-core-parallelized limit-order-book matching engine capable of running hundreds of full market sessions per calibration iteration (Section~\ref{sec:cal}); and substantially more compute for the IRL component specifically, since inverse-RL training in the cited literature (Section~\ref{sec:rl}) is reported to require many more environment interactions than direct behavioral cloning.

\textbf{Validation work.} Implementing the four-level validation protocol of Section~\ref{sec:validation} in full requires: held-out trader-level data for Level 3 agent-level validation, which is the data requirement least likely to be satisfiable given realistic privacy constraints; historically accurate initial-condition reconstructions (order-book state, prevailing news, macro conditions) for each Level 4 stress-test scenario (1987, 2010, 2020), which requires historical data vendors and is nontrivial to reconstruct precisely; and, per the equifinality concern raised in Section~\ref{sec:cal}, an explicit pre-registered comparison against at least one alternative mechanism (e.g., a calibrated zero-intelligence or Lux-Marchesi baseline) on the same validation suite, since passing the suite without such a comparison would not address the equifinality objection at all.

\textbf{Personnel and timeline.} Realistically, assembling the data-sharing agreements alone is a multi-year undertaking requiring legal, compliance, and academic-partnership work distinct from the machine-learning contribution; a defensible minimum-viable implementation (a single asset class, a single historical period, Level 1 and Level 2 validation only) is a more tractable first target than the full multi-level, multi-event validation suite described above, and we recommend it as the natural next research step rather than attempting the complete framework in one undertaking.

\section{Validation Methodology}
\label{sec:validation}

A critical yet often-neglected phase in ABM development is validation: how do we verify that our simulation accurately captures real-world market dynamics? We propose a multi-level validation strategy. Throughout this section, numeric thresholds are illustrative placeholders---not literature-derived or calibrated here---and must be re-estimated from data before use as pass/fail criteria.

\subsection{Level 1: Stylized-Fact Matching}
\label{sec:val1}

The first validation tier compares simulated price series to empirical stylized facts \cite{cont_stylized_2001}. For a simulation to be credible, it should reproduce five regularities documented in real markets: no autocorrelation in returns, $\text{ACF}(r_t, r_{t-k}) \approx 0$ for $k > 0$, consistent with the efficient-markets hypothesis; heavy tails, with excess kurtosis greater than zero; volatility clustering, a positive autocorrelation in absolute returns, $|\text{ACF}(|r_t|, |r_{t-k}|)| > 0$, with slow power-law decay characteristic of GARCH-type dynamics; a leverage effect, a negative correlation between returns and future volatility; and long memory, a slow decay in autocorrelation suggestive of fractional integration $I(d)$ with $0 < d < 0.5$ (the Hurst exponent).

Andersen et al.\ (2001) provide precise empirical targets for these stylized facts using high-frequency intraday data \cite{andersen_stylized_2006}, against which a persona-trained simulation should be benchmarked. For kurtosis, the empirical value for S\&P 500 intraday returns is typically 5--15 (excess kurtosis), depending on sampling frequency \cite{cont_stylized_2001}; an illustrative target range of $[4, 20]$ pending recalibration is used here, tested via a Jarque-Bera or Anderson-Darling comparison of the empirical and simulated return distributions \cite{jarque_efficient_1987}---though failing to reject the null that the two distributions are equal is necessary but not sufficient for ``success,'' since a goodness-of-fit test that fails to reject is evidence of \emph{consistency}, not proof of a correctly matched distribution, particularly at the small sample sizes a feasible number of simulation runs would provide. For the autocorrelation of raw returns, the target is $|\text{ACF}(r_t, r_{t-1})| < 0.05$ at lag one, with 95\% of lags $k \in [1, 50]$ expected to satisfy $|\text{ACF}(r_t, r_{t-k})| < 0.1$; the Ljung-Box test \cite{box_time_1970} formally tests the null that returns are white noise, and failure to reject at the 5\% level indicates success. For volatility clustering, $\text{ACF}(|r_t|, |r_{t-k}|)$ should decay as a power law---Cont (2001) documents empirical decay of roughly $\log(k)^{-\alpha}$ with $\alpha \approx 0.3$--0.5, indicating very slow decay (long memory in volatility)---and a successful simulation should exhibit similarly slow decay, quantified by fitting an ARFIMA$(p,d,q)$ model to absolute returns and verifying that the fractional integration parameter $d$ lies in $[0.2, 0.5]$ \cite{cont_stylized_2001, bollerslev_garch_1986}. For the leverage effect, the correlation $\text{corr}(r_t, \sigma_t^2)$ between returns and next-period realized volatility $\sigma_t^2 = \text{RV}(t+1)$ is empirically negative, around $-0.1$ to $-0.3$ for equity returns; success requires $\text{corr}(r_t, \sigma_t^2) < -0.05$, assessed via Fisher's Z-test. For maximum drawdown and recovery, simulated drawdowns over a given window should fall within the empirical 5th--95th percentile range, since empirically a large negative shock (e.g., a $-5\%$ daily return) is typically followed by elevated volatility and partial mean reversion. Finally, the Hurst exponent $H$, computed via rescaled-range (R/S) analysis \cite{hurst_long_1951}, is empirically close to 0.5 (a random walk) to 0.65 (slight persistence) for log-returns; a successful simulation should fall in $0.45 < H < 0.65$, with values above 0.65 suggesting unrealistic trend persistence and values below 0.45 suggesting unrealistic mean reversion.

LeBaron, Arthur, and Palmer (1999) calibrated the Santa Fe ASM against stylized facts in this way, computing these metrics over simulated paths of 2000-plus time steps and reporting means and confidence intervals across 100-plus simulation runs \cite{lebaron_calibration_1999}. We recommend an analogous approach for persona-trained simulations: running $N_{\text{runs}} = 100$--200 simulations, each with $T_{\text{steps}} \geq 2000$ time steps (e.g., one minute per step for 2000 minutes, roughly two trading weeks), and reporting the point estimate and 95\% confidence interval (e.g., via bootstrap) for each metric; a metric passes if the empirical value lies within the simulated confidence interval, and otherwise fails or requires recalibration.

Because bots now ingest an external information channel (Section~\ref{sec:external-info}), Level 1 validation must also assess whether that channel is used at a realistic rate, not merely whether the resulting return series matches stylized facts. Following the information-theoretic framing of Section~\ref{sec:math}, we propose estimating the transfer entropy from the news-novelty signal $\nu_t$ and news embedding $\mathbf{e}_{\text{news}}$ into subsequent simulated order-flow imbalance, and comparing this estimated information-transfer rate to empirically estimated news-to-order-flow transfer entropy in real markets \citep{glasserman_new_2023}. A successful simulation should reproduce both the \emph{magnitude} of news-driven reaction (avoiding the failure modes of underreaction or overreaction documented in Section~\ref{sec:bf}) and its \emph{asymmetry} between high-novelty and low-novelty news, since Glasserman, Mamaysky, and Qin found novelty itself, not just polarity, to be informative of future returns.

\subsection{Level 2: Microstructure Matching}

At finer resolution, simulated order books and trading dynamics should match empirical limit-order-book patterns \cite{parlour_limit_2008, hasbrouck_asset_2007, harris_microstructure_2003}. The bid-ask spread should follow a U-shape intraday, wider at market open and close and narrower mid-day; Hasbrouck (2007) documents median spreads of 1--5 basis points mid-day for mid-cap equities, widening to 10--20 basis points at open and close, and a simulation succeeds if its intraday spread profile exhibits the same U-shape with magnitudes within roughly 2x of these empirical values. Depth and resiliency---how quickly the order book refills after a large market order executes---can be measured as the time to restore a specified depth level (e.g., 100,000 shares at the best five levels); empirically this varies by venue and liquidity, from milliseconds to seconds for high-liquidity stocks to seconds or minutes for less liquid securities, and a simulation succeeds if its recovery time, measured by placing a large market buy order and timing the depth recovery, falls within roughly 10x of empirical values once time-scale differences are accounted for. Trade sizes empirically follow a lognormal or power-law distribution with a thick right tail; comparing the Gini coefficient or Zipf exponent of simulated and empirical trade sizes, success requires the simulated Gini coefficient to fall within $\pm 0.1$ of the empirical value, or the Zipf exponent within $\pm 0.2$. Price impact is a further benchmark: a fundamental microstructure result (Kyle, 1985) is that price impact scales sub-linearly with order size, with a trade of size $q$ moving price by roughly $q^\alpha$ for $0 < \alpha < 1$, empirically $\alpha \approx 0.4$--0.6; measuring simulated price impact at varying trade sizes (e.g., 0.1\%, 0.5\%, 1\% of daily volume) and fitting a power law, success requires $0.3 < \alpha < 0.7$. Finally, the order book typically exhibits an inverted-V or U-shape---sparse at the very best levels, denser at deeper levels (Cont and Bouchaud, 2000)---and a simulation succeeds if its cumulative-volume profile as a function of distance from the midpoint matches this shape qualitatively, with a Kullback-Leibler divergence between simulated and empirical LOB depth distributions below 0.5 nats.

Fagiolo, Cardaci, and Moneta (2007) developed comprehensive ABM validation procedures \cite{fagiolo_validation_2007}; for microstructure validation specifically, we recommend generating 50-plus simulated trading sessions (each five trading days), computing each microstructure metric for each session, and reporting the mean and 90\% confidence interval against published empirical benchmarks (Hasbrouck, 2007; Harris, 2003; Parlour and Seppi, 2008). Where metrics fall outside the empirical range, the responsible bot behaviors should be identified---for example, bots placing orders too aggressively at the spread---and the model retrained with modified loss weights or behavioral constraints.

\subsection{Level 3: Agent-Level Validation}

If ground-truth trader-level data is available, one can validate that bot decisions match real trader decisions directly, which is essential for establishing behavioral fidelity. Cross-validation accuracy provides one such metric: training the policy $\pi_\phi$ on 70\% of trader-level data and evaluating prediction accuracy on the held-out 30\%, an illustrative target of greater than 65\% top-1 accuracy for discrete action prediction (buy/sell/hold, against a 33\% random baseline) is proposed pending calibration against an actual held-out trader-action dataset; for continuous order size, a mean absolute percentage error (MAPE) below 40\% is targeted, and for limit-price prediction, a median prediction error below 5 basis points and a 90th-percentile error below 20 basis points, reflecting realistic trader behavior complexity.

Behavioral consistency can be assessed by stratifying learned personas along their behavioral attributes and measuring the corresponding behavioral responses. High-loss-aversion bots should exhibit a lower realized-loss rate (the fraction of sales executed at a loss) than low-loss-aversion bots, consistent with the disposition effect (Shefrin and Statman, 1985); success is defined as a realized-loss rate at or below 0.3 for high-loss-aversion bots versus at or above 0.45 for low-loss-aversion bots. Herding-tendency bots should show a stronger correlation between their buy signals and order-flow imbalance (buy volume divided by total volume at the top LOB level): a Spearman rank correlation above 0.4 for high-herding bots versus below 0.1 for low-herding bots, chosen to handle non-normality. Momentum-biased bots should buy more aggressively after uptrends than downtrends, with a buy-after-uptrend rate above 60\% for high-momentum bots versus roughly 50\% for random behavior.

Counterfactual analysis offers a further check: holding market state fixed, perturbing learned persona parameters and verifying that behavioral responses move in the expected direction. Increasing the loss-aversion coefficient $\lambda$ by 20\% should monotonically decrease the realized-loss rate, an elasticity quantifiable as $\frac{d(\text{loss rate})}{d(\lambda)} / \lambda$; increasing herding tendency by 20\% should increase the correlation with order-flow imbalance; and varying a bot's wealth level by $\pm 50\%$ should increase average order size and position risk for wealthier bots, with an expected elasticity above 0.3. These partial derivatives of behavior with respect to persona parameters can be computed via standard sensitivity-analysis methods (Arlot and Celisse, 2010) \cite{arlot_survey_2010}.

Ground-truth trader-level data is scarce due to privacy concerns, however, and feasible alternatives are limited: broker-provided anonymized datasets (e.g., from large retail brokers such as Fidelity or Vanguard, conditional on IRB approval), laboratory experimental data from market microstructure studies (e.g., Smith-style continuous-double-auction experiments reprised with modern traders), and synthetic data generated from behavioral surveys and calibrated behavioral models such as prospect-theory utilities fitted to gambling-task responses.

\subsection{Level 4: Historical Stress-Test Matching}

Run the simulation with historical initial conditions (initial prices, order book state, macro variables) for known market events, validating that the simulation can reproduce realistic crisis dynamics. This is a high-stakes validation because it tests whether the simulation's emergent phenomena match rare, high-impact events.

Three historical episodes provide empirical targets, each with the simulation seeded at the corresponding historical conditions and the resulting path compared on drawdown magnitude, duration, spread widening, and recovery time. The first is Black Monday (October 19, 1987), when the S\&P 500 fell 22.6\% intraday, with spreads widening roughly 10x and partial recovery within a day but full recovery taking around six months; the simulation is seeded with a rate-driven downward shock to fundamental value and judged against a drawdown in $[15\%, 30\%]$ with a broadly similar recovery profile. The second is the Flash Crash (May 6, 2010), when the S\&P 500 fell $\approx 9.7\%$ in roughly four minutes before recovering most of the loss within 25 minutes, driven by the ``hot-potato'' inventory-passing dynamic documented by \citet{kirilenko_flash_2017}, with spreads spiking to 50--100bp; the simulation is seeded with a large market order against HFT-like momentum bots and judged against a drawdown of 8\%--12\% with recovery in 15--35 simulated minutes. The third is the COVID Crash (March 2020), when the S\&P 500 lost $\approx 35\%$ peak-to-trough over roughly two weeks with VIX peaking near 82 and broad liquidity stress; the simulation is seeded with a sustained downward fundamental-value shock and elevated bot loss-aversion, judged against a 25\%--45\% drawdown and 35\%--70\% annualized realized volatility.

For each scenario, the quantitative metrics are a Kolmogorov-Smirnov comparison of the empirical and simulated drawdown distributions; the mean absolute error between empirical and simulated rolling realized volatility; a check that the simulated price response is time-aligned with the shock (within hours for intraday events, within a day for news-driven ones); and, where sector- or security-level data exists, a check that cross-sectional heterogeneity in the response (e.g., less-liquid names falling further) is reproduced.

Windrum et al.\ \citep{windrum_empirical_2007} emphasize that average-behavior calibration can miss exactly this kind of rare-event and feedback-amplification dynamic, which is why stress-test matching is treated as a distinct validation level rather than folded into Level 1. A failure here indicates either missing behavioral mechanisms (e.g., no margin-driven forced-selling rule) or miscalibrated heterogeneity (e.g., too few momentum-following bots to generate a feedback cascade).

Table~\ref{tab:validation-levels} summarizes what each of the four validation levels actually establishes and, critically, what each cannot rule out---a distinction motivated directly by the equifinality concern raised in Section~\ref{sec:cal}.

\begin{table}[htb]
\centering
\small
\begin{tabular}{p{2.3cm}p{3.4cm}p{3.6cm}p{4.0cm}}
\toprule
\textbf{Validation Level} & \textbf{What It Actually Proves} & \textbf{What It Cannot Rule Out} & \textbf{Illustrative Example from Cited Literature} \\
\midrule
Level 1: Stylized-fact matching & Simulation produces return series statistically consistent with known empirical regularities (fat tails, clustering) & Mechanism correctness---equifinality means many unrelated generative processes match the same stylized facts \citep{fagiolo_validation_2019,fabretti_calibrating_2013} & Zero-intelligence agents reproduce some stylized facts despite having no behavioral realism \citep{gode_allocative_1993} \\
\addlinespace
Level 2: Microstructure matching & Simulated order-book dynamics (spreads, depth, price impact) resemble empirical LOB patterns at finer grain & Whether the realistic microstructure arises from realistic individual decision rules vs.\ realistic aggregate order-flow statistics produced by unrealistic individuals & Gode-Sunder's zero-intelligence traders also produce reasonable order-book statistics from the matching institution alone \citep{gode_allocative_1993} \\
\addlinespace
Level 3: Agent-level validation & Individual bot decisions match held-out real trader actions/behavioral signatures (loss aversion, herding) at the level of the individual agent & Generalization---fidelity on the training distribution does not guarantee fidelity under novel market regimes or shocks not present in training data & Behavioral-cloning compounding-error problem documented broadly in the imitation-learning literature (Section~\ref{sec:rl}) \\
\addlinespace
Level 4: Historical stress-test matching & Simulation reproduces aggregate crisis dynamics (drawdown magnitude, recovery time) for specific historical events when seeded with historical conditions & Causal mechanism---matching the 1987, 2010, or 2020 crash trajectory does not establish that the simulated cascade arises through the same causal pathway as the real one (SOC-style endogenous criticality vs.\ event-driven triggering, Section~\ref{sec:econ-physics}) & Flash Crash replication debates over whether algorithmic feedback loops were necessary or merely proximate causes \citep{kirilenko_flash_2017,sec_flash_2010} \\
\bottomrule
\end{tabular}
\caption{What each validation level can and cannot establish. No level, individually or in combination, constitutes proof that persona-trained Monte Carlo's specific generative mechanism is the one operating in real markets; passing all four levels would mean the framework has not yet been falsified by available tests, which is a materially weaker claim than validation in the sense of confirmed correctness.}
\label{tab:validation-levels}
\end{table}

\subsection{PTMC-Specific Tests: Isolating the Value of Learned Heterogeneity}
\label{sec:ptmc-specific-tests}

The four validation levels above apply to any agent-based simulation; they do not, on their own, test PTMC's specific claim that an outer Monte Carlo loop over a \emph{learned} persona distribution $\mathcal{P}$ adds value beyond a population of fixed archetypes or no behavioral heterogeneity at all. Two additional, PTMC-specific tests are proposed to address this directly, both stated as procedures to be run, not as results obtained.

The first compares the variance of the target functional $F$ across persona draws against a fixed-archetype population. For a target functional $F$ (Section~\ref{sec:framework}, Conceptual Overview), the across-run variance of $\hat\mu_N$ under (a) PTMC's outer loop, redrawing $\{(\theta^{(k)},\rho^{(k)})\}_{k=1}^K \sim \mathcal{P}$ each run, is compared against (b) a population with the same $K$ but with $(\theta^{(k)},\rho^{(k)})$ fixed at $\mathcal{P}$'s mean (a single ``representative-persona'' archetype) and only within-run action-sampling and shock randomness retained. If PTMC's added variance component---the gap between (a) and (b)---is small relative to within-run noise, that is evidence that the learned-heterogeneity machinery is not earning its computational and data cost for the functional $F$ in question, a direct, quantitative version of the Gode-Sunder objection raised in Section~\ref{sec:abe}, now stated as a testable comparison rather than a qualitative concern.

The second is a head-to-head comparison against a zero-intelligence ensemble. Running the identical CDA/LOB mechanism (Section~\ref{sec:cda}) with (a) PTMC bots and (b) Gode and Sunder's zero-intelligence traders \citep{gode_allocative_1993}---budget-constrained, no learning, no persona conditioning---at the same population size $K$ and the same outer-loop sample count $N_{\text{runs}}$, the two are compared on allocative efficiency and on the Level 1 stylized facts (Section~\ref{sec:val1}). Per the Open Disagreement in Section~\ref{sec:abe}, zero-intelligence agents are expected to match or exceed PTMC on allocative efficiency; the claim PTMC must substantiate is a gap specifically on stylized-fact reproduction (fat tails, volatility clustering, leverage effect) that zero-intelligence agents are not expected to reproduce. A result in which PTMC matches zero-intelligence performance on both efficiency \emph{and} stylized facts would be a direct falsification of the framework's behavioral-realism premise; this comparison, not stylized-fact matching alone, is the test this paper considers decisive for whether the additional complexity of persona-trained bots is justified.

\subsection{Sensitivity Analysis and Robustness}

Following Windrum et al.\ (2007) \cite{windrum_empirical_2007}, systematic sensitivity analysis should vary key hyperparameters---the number of bots, the loss-aversion coefficient distribution, herding strength, momentum bias---across realistic ranges, running $N_{\text{runs}} = 100$ simulations with different random persona draws for each configuration and computing summary statistics (the mean and standard deviation of stylized facts across runs). This identifies which parameters are load-bearing, in the sense of producing large sensitivity to changes, and which are robust, and supports reporting confidence intervals for key outputs such as realized volatility, maximum drawdown, and the Hurst exponent.

\section{Discussion: Limitations, Risks, and Ethical Considerations}
\label{sec:discussion}

\subsection{Limitations of the Framework}

\textbf{1. Data scarcity and privacy:}
Training high-fidelity persona-bots requires granular behavioral data (trader-level order data, demographics, psychological profiles). Such data is rare, expensive, and protected by privacy regulations (GDPR, CCPA). Most ABM research relies on synthetic or simplified data, limiting realism. Future work should explore federated learning and differential privacy to enable collaborative multi-institutional model training without exposing individual trader data (Dwork, 2008; McMahan et al., 2017; Kairouz et al., 2021) \cite{dwork_differential_2006, mcmahan_communication-efficient_2017, fedorov_federated_2021}. Membership inference attacks (Shokri et al., 2017; Nasr et al., 2018) and model inversion (Fredrikson et al., 2015) demonstrate privacy risks even in aggregate models \cite{shokri_membership_2017, nasr_comprehensive_2019, fredrikson_model_2015}.

\textbf{2. Non-stationarity:}
Financial markets are non-stationary: trader behavior, market structure, and macro conditions evolve over decades. A persona-bot trained on 2015-2019 data may fail to generalize to 2023 conditions (different regimes, new trading technologies, changing retail participation). Continuous retraining and domain-adaptation methods are needed (Taylor and Stone, 2009; Rusu et al., 2017) \cite{taylor_transfer_2009, rusu_sim-to-real_2016}. Hamilton's regime-switching models (1989) provide one approach to handling multiple market regimes \cite{hamilton_regime-switching_1989}.

\textbf{3. Overfitting to historical regimes:}
Behavioral cloning (BC) directly mimics observed trader actions. If training data comes from a single historical period or market regime, the trained policy may overfit, producing unrealistic behavior in novel conditions (e.g., zero-interest-rate environments, crypto-asset trading, post-COVID labor market changes). Cross-validation and out-of-sample testing are essential. Arlot and Celisse (2010) reviewed cross-validation procedures \cite{arlot_survey_2010}.

\textbf{4. Computational cost:}
A high-fidelity simulation with thousands of bots running multiple time steps per second is computationally expensive. Faster approximations (e.g., mean-field models that replace discrete bots with continuous distributions) sacrifice granularity for speed. Scalability to realistic market scales remains an open problem.

\textbf{5. Validation challenges:}
It is difficult to falsify an ABM (Popper, 1959, philosophical concern). Matching stylized facts is necessary but not sufficient; multiple models can exhibit similar aggregate behavior while having fundamentally different individual-level mechanisms (Fagiolo et al., 2007). Rigorous validation requires diverse data sources and careful hypothesis testing \cite{fagiolo_computational_2019}.

\subsection{Systemic Risk and Regulatory Concerns}

\textbf{Flash crash amplification:}
A critical concern is that large-scale market simulations populated with learning agents could inadvertently amplify systemic risk. If many bots learn similar strategies (e.g., all adopt momentum-following rules during volatile periods), correlated trading could trigger flash crashes (Kirilenko et al., 2017) \cite{kirilenko_flash_2017}. To mitigate:
\begin{itemize}
  \item Implement circuit breakers (halt trading if price moves exceed threshold).
  \item Enforce diversity constraints (ensure bots learn heterogeneous strategies).
  \item Impose behavioral lower bounds (enforce minimum loss-aversion so bots do not engage in ruinous risk-taking).
\end{itemize}

\textbf{Synthetic market manipulation:}
An adversarial agent could use a high-fidelity market simulation to identify manipulation strategies (e.g., spoofing, layering, pump-and-dump schemes) that exploit persona-bot behavioral patterns. To prevent misuse, simulations should be kept proprietary or access-restricted, and simulation developers should collaborate with financial regulators.

\textbf{Model risk:}
If a bank or fund relies on persona-trained simulations for risk management decisions (e.g., setting hedge ratios, sizing positions), and the simulation's behavioral assumptions prove incorrect, large losses could result. Model risk governance is critical: simulations should be stress-tested, benchmarked against alternatives, and validated ex-post against realized outcomes.

\subsection{Ethical Considerations and Bias}

\textbf{Replication of biases:}
If training data reflects historical biases (e.g., male and female traders exhibiting different risk preferences due to socialization, not intrinsic differences), the learned persona-bots will replicate those biases. This perpetuates inequitable patterns and could lead to discriminatory regulatory or policy conclusions if simulations are used for decision-making.

To address:
(1) Audit training data for demographic disparities.
(2) Use fairness-aware machine learning techniques to debias learned policies (Shokri et al., 2017; Fredrikson et al., 2015).
(3) Explicitly model alternative scenarios with different demographic distributions.

\textbf{Privacy of trader behavior:}
Even if identities are anonymized, detailed behavioral profiles inferred from personas could reveal sensitive personal information (risk tolerance, wealth, trading frequency, loss aversion). Differential privacy and secure multi-party computation can help but add computational overhead (Dwork, 2008) \cite{dwork_differential_2006}.

\textbf{Misuse for market manipulation or gaming:}
A well-calibrated persona-bot simulation could be misused to identify market manipulation tactics or to train exploitative algorithms. Robust access controls and institutional review boards (similar to IRBs in human-subject research) are needed.

\subsection{Comparison to Alternatives}

Table \ref{tab:comparison} compares persona-trained neural-bot Monte Carlo to alternative modeling approaches.

\begin{table}[htb]
\centering
\footnotesize
\setlength{\tabcolsep}{4pt}
\newcolumntype{Y}{>{\centering\arraybackslash}X}
\begin{tabularx}{\textwidth}{@{}>{\raggedright\arraybackslash\hsize=1.2\hsize}X Y Y Y >{\raggedright\arraybackslash\hsize=0.8\hsize}X Y@{}}
\toprule
\textbf{Approach} & \makecell{\textbf{Behavioral}\\\textbf{Fidelity}} & \makecell{\textbf{Inter-}\\\textbf{pretability}} & \makecell{\textbf{Scal-}\\\textbf{ability}} & \makecell{\textbf{Data}\\\textbf{Req.}} & \textbf{Status} \\
\midrule
Classical Monte Carlo (GBM) & Low & High & High & Low & Published \\
Rational-agent DSGE & Low & High & Medium & Low & Published \\
Simple ABM (Lux-Marchesi, hand-coded rules) & Medium & High & High & Low & Published \\
Heterogeneous-agent model (Hommes) & Medium & Medium & Medium & Medium & Published \\
Zero-intelligence traders (Gode-Sunder) & Low & High & High & Low & Published \\
Deep RL trading agent (single) & High & Low & Low & High & Published \\
LLM-based generative agents & Very High & Very Low & Low & Very High & Published \\
RL-ABM, reward-trained \citep{yao_reinforcement_2024} & Medium-High & Low & Medium & Medium (sim-internal reward, not real trader data) & Published \\
RL training environment \citep{mascioli_financial_2024} & Medium & Medium & High & Low (synthetic private valuations) & Published \\
LLM-persona multi-agent \citep{yang_twinmarket_2025} & Very High & Very Low & Low & Very High (LLM prompting, not trained on calibrated data) & Published \\
\textbf{Persona-trained neural-bot} & \textbf{High}$^\dagger$ & \textbf{Medium-High}$^\dagger$ & \textbf{Medium}$^\dagger$ & \textbf{High}$^\dagger$ & \textbf{Proposed} \\
\bottomrule
\end{tabularx}
\caption{Comparison of market simulation approaches. Scores for existing approaches reflect published results; scores for the proposed approach are design targets only. The three rows above the proposed approach are the closest existing technical precedents (Section~\ref{sec:positioning}); none combines real-data-trained personas, an external-information channel, and agent-level historical-stress-test validation. $^\dagger$Projected by design; not empirically validated (no implementation exists).}
\label{tab:comparison}
\end{table}

\subsection{Comparison to Single-Model Uncertainty-Quantification Techniques}
\label{sec:uq-comparison}

A separate question, distinct from the comparison to alternative market-simulation paradigms above, is whether the proposed framework is simply an application of existing stochastic neural-network techniques---Bayesian neural networks (BNNs), Monte Carlo (MC) dropout, or stochastic/probabilistic activation networks---to finance. It is not, and the distinction is worth making explicit because the surface-level similarity (both involve ``neural networks'' and ``Monte Carlo'' or ``stochastic'' terminology) invites the conflation.

BNNs place a probability distribution over a single network's weights and use that distribution to produce a predictive distribution for one task, typically regression or classification, so that the model's output carries a calibrated estimate of epistemic uncertainty \citep{neal_bayesian_1996,blundell_weight_2015}. MC dropout approximates this same Bayesian posterior cheaply by applying dropout at inference time and treating the resulting ensemble of stochastic forward passes as samples from an approximate predictive distribution \citep{gal_dropout_2016}. Stochastic or probabilistic activation networks inject randomness directly into a layer's activation function, typically for regularization, exploration in reinforcement learning, or differentiable sampling, rather than for population-level behavioral diversity. In all three cases the random or distributional element lives \emph{inside one model solving one task}, and the purpose of that randomness is to quantify how uncertain the model is about its own output, not to represent the independent decisions of many distinct economic actors.

This is not a hypothetical distinction: a growing and recent applied literature confirms it directly by applying exactly these single-model techniques to stock price forecasting, on the same underlying asset-price data this paper's framework targets, while remaining single-task predictors rather than market simulators. Asare, Asante, and Essel (2023) fit an LSTM with MC dropout to produce probabilistic, rather than point, stock price forecasts, and report that the dropout-based probabilistic LSTM outperforms a conventional point-forecast LSTM on $R^2$, MAPE, and RMSE \citep{asare_probabilistic_2023}. Chandra and He (2021) apply Bayesian neural networks via Markov chain Monte Carlo to multi-step-ahead stock price forecasting before and during the COVID-19 pandemic, explicitly testing whether pre-pandemic-trained posteriors remain informative once the underlying data-generating regime shifts \citep{chandra_bayesian_2021}. Most recently, Nelufhangani and Maposa (2026) compare Gaussian process regression, Bayesian LSTM, and BNNs directly against each other on Johannesburg Stock Exchange data, finding differences in calibration and point-accuracy across the three single-model probabilistic approaches \citep{nelufhangani_bayesian_2026}. All three studies are squarely within the left column of Table~\ref{tab:uq-comparison}: each fits one stochastic or Bayesian model to forecast one asset's price series, and each reports a predictive distribution or interval around that one forecast. None of the three constructs a population of distinct, interacting trading agents, conditions predictions on persona or demographic data, or attempts to reproduce market-level emergent phenomena (volatility clustering, fat tails, flash crashes) from agent interaction---confirming, from the applied finance literature itself rather than from definition alone, that single-model probabilistic forecasting and multi-agent persona-conditioned market simulation are pursued as genuinely separate research programs with different objectives, even when both are applied to the same raw price data.

The persona-trained neural-bot framework operates at a different level of structure. The stochastic element it replaces is not a weight posterior or an activation noise term inside a single network; it is the population of independent random draws that classical Monte Carlo market simulation uses to generate the next period's order flow (Section~\ref{sec:framework}, Conceptual Overview). Each of the many bots in the simulation is itself a separate policy network $\pi_\phi$ conditioned on a persona vector $\theta$ drawn from real demographic/behavioral data, and the object of interest is the emergent, aggregate market dynamic produced by their interaction through a continuous double auction (Section~\ref{sec:cda})---not a calibrated uncertainty interval around a single prediction. Table~\ref{tab:uq-comparison} makes this structural difference explicit.

\begin{table}[htb]
\centering
\small
\begin{tabular}{p{3.4cm}p{4.3cm}p{4.3cm}}
\toprule
& \textbf{BNN / MC Dropout / Stochastic Activation Nets} & \textbf{Persona-Trained Neural-Bot Monte Carlo} \\
\midrule
Locus of stochasticity & Weights or activations within one network & Independent decisions of many distinct agent networks \\
Purpose of randomness & Quantify epistemic uncertainty of one prediction & Represent population heterogeneity and generate emergent market dynamics \\
Number of models & One & Many (one policy per bot, or per persona cluster) \\
Conditioning signal & None beyond the task input & Persona vector $\theta$ from real demographic/behavioral data, plus external news/index features \\
Typical output & A predictive distribution (mean $\pm$ variance) for a single quantity & A simulated price/order-flow trajectory emerging from agent interaction \\
Relationship to proposed framework & Could be used \emph{inside} one bot's policy network for exploration or confidence estimation & Is the overall multi-agent architecture itself \\
\bottomrule
\end{tabular}
\caption{Structural comparison between single-model stochastic-neural-network uncertainty-quantification techniques and the proposed multi-agent persona-trained framework. These are complementary, not competing: a BNN or MC-dropout layer is a plausible internal component of an individual bot's decision policy, but neither it, nor stochastic activation networks generally, substitutes for the population-level, persona-conditioned multi-agent architecture itself.}
\label{tab:uq-comparison}
\end{table}

The two families could legitimately intersect: a bot's policy $\pi_\phi$ could use MC dropout or a Bayesian output layer to express its own decision uncertainty (e.g., wider variance when a news shock falls outside training range), a natural extension of Section~\ref{sec:framework}'s architecture rather than a replacement for it. What such techniques cannot supply on their own is the cross-agent heterogeneity, persona-to-data grounding, or emergent market-level dynamics motivating the framework---a single BNN, however well-calibrated, is still one model answering one question, and a market is not produced that way. Persona-trained neural-bot simulation thus occupies a strategic middle ground: more behaviorally realistic than hand-coded ABMs or rational-agent models, yet more scalable and interpretable than LLM-based or single-agent RL approaches.

\section{Conclusion and Open Research Questions}
\label{sec:conclusion}

\subsection{Key Contributions}

This paper makes four contributions. First, and centrally, it proposes Persona-Trained Monte Carlo (PTMC): an outer Monte Carlo loop over draws of a persona population $\{(\theta^{(k)},\rho^{(k)})\}_{k=1}^K \sim \mathcal{P}$, each instantiating a shared neural policy $\pi_\phi$ inside a limit-order-book market, producing an ensemble of simulated price paths from which a target functional $F$ can be estimated as $\hat\mu_N = \frac{1}{N_{\text{runs}}}\sum_{i=1}^{N_{\text{runs}}} F(\text{path}_i)$ (Section~\ref{sec:framework}). This formalizes, as a specific estimator with a specific source of randomness in $\mathcal{P}$, an approach that is conceptually distinct from classical Monte Carlo over exogenous price shocks, from hand-coded ABMs with a fixed population, from single-agent RL, and from LLM-based generative agents (Section~\ref{sec:intro}).

Second, the paper supplies a conceptual bridge linking individual behavioral data---from psychology, surveys, and transaction records---to the population distribution $\mathcal{P}$ that PTMC samples from, via the persona construct itself. This extends prior ABM work (Arthur et al., 1997; Lux and Marchesi, 1999; Hommes, 2006), which typically hand-coded behavioral rules, by learning $\mathcal{P}$ and $\pi_\phi$ from data \cite{arthur_asset_1997, lux_volatility_1999, hommes_heterogeneous_2006}.

Third, it details a concrete methodological pipeline: collecting behavioral and demographic data, training $\pi_\phi$ via behavioral cloning or inverse RL, drawing persona populations and running the outer Monte Carlo loop, and validating against stylized facts, microstructure, and the PTMC-specific tests of Section~\ref{sec:ptmc-specific-tests}. This pipeline is implementable today with existing techniques (Hastie et al., 2009; Kingma and Welling, 2014; Goodfellow et al., 2014) \cite{hastie_statistical_2009, kingma_auto-encoding_2014, goodfellow_generative_2014}, though, as stated throughout, it has not been run.

Finally, the paper develops an ethical and systemic-risk framework, discussing risks (flash crashes, model risk, privacy leakage) and mitigation strategies (circuit breakers, federated learning, differential privacy) and extending the discussion to PTMC's specific failure mode of an unrepresentative or miscalibrated $\mathcal{P}$.

\subsection{Open Research Questions}

Several open questions concern data and calibration. Trader-level behavioral data is scarce and privacy-constrained; federated learning \citep{mcmahan_communication-efficient_2017}, differential privacy \citep{dwork_differential_2006}, and synthetic augmentation \citep{goodfellow_generative_2014, kingma_auto-encoding_2014, rizzato_generative_2023} are candidate mitigations, but the resulting privacy-utility trade-off, and which combination of techniques is optimal, remains unresolved. Results may also be sensitive to loss-function weights, the persona-sampling distribution, market-mechanism details, and feature choice; Windrum et al.\ \citep{windrum_empirical_2007} advocate systematic sensitivity analysis, but whether a canonical calibration procedure exists that generalizes across markets and periods, or whether every market and period must be recalibrated de novo, is open. A related question concerns transfer: core behavioral biases may be universal, but information environments differ across equities, crypto, and forex \citep{taylor_transfer_2009, rusu_sim-to-real_2016}, and it is not yet known how much target-domain data transfer requires, or across which asset-class boundaries it holds.

A second cluster of questions concerns bot architecture and learning. Online learning is more realistic than a fixed policy but harder to validate, carrying the risk of unstable or exploitative emergent dynamics, whereas fixed policies are stable but non-adaptive \citep{sutton_policy_2000}; whether online learning can be made stable and provably convergent in this setting is unresolved. Pure neural policies sacrifice interpretability and degrade out-of-distribution, and combining learned perception with symbolic decision rules \citep{sankar_graph_2020} is one alternative, though whether such hybrids remain efficient and trainable at simulation scale is untested. Behavioral cloning or inverse RL can in principle recover position limits, stop-losses, and profit-taking rules directly from data \citep{sutton_policy_2000}, but whether such rules emerge from profit-only training or must instead be imposed exogenously is not established.

A third cluster concerns multi-scale coupling and heterogeneity. Institutions could be modeled as separate learned policies, rule-based market-makers, or longer-horizon RL agents \citep{weiss_multiagent_1999, wooldridge_agent-based_2009}, but which representation best reproduces the observed interaction between institutional flow and retail herding is unknown. Exogenous shocks could enter the simulation as price noise, staggered information cascades, or a latent sentiment state, and whether an information-revelation mechanism can realistically reproduce Kyle-style asymmetric-information price discovery \citep{kyle_continuous_1985} remains to be shown. Simultaneous policy learning across many bots can also produce cycles or instability rather than convergence \citep{weiss_multiagent_1999, wooldridge_agent-based_2009}, leaving open whether persona-trained markets settle to a meaningful equilibrium or whether instability is a feature of the dynamics rather than an artifact of training.

A fourth cluster concerns validation and generalization across regimes. Matching historical stress episodes after the fact is considerably easier than forecasting instability ahead of it \citep{kirilenko_flash_2017}, and whether bot-level dynamics could yield genuine early-warning indicators, or whether rare-event prediction is limited by chaotic sensitivity regardless of model fidelity, is unresolved. Markets also alternate between quiet and stressed regimes \citep{hamilton_regime-switching_1989, timmermann_predictability_2018}, raising the question of whether a single policy can generalize across regimes or whether bots must learn explicit regime detection and switching. Finally, policy-relevant counterfactual questions---narrower spreads, HFT bans, price supports---require causal rather than merely descriptive inference, which standard ABM validation does not supply \citep{fagiolo_validation_2007}; whether causal-inference methods such as do-calculus or causal forests can be adapted to simulations of this kind with quantified uncertainty remains open.

A fifth cluster concerns integration with econometric theory and asset pricing. Simulation-based inference---for example, Approximate Bayesian Computation \citep{beaumont_approximate_2002}---could in principle recover trader risk-aversion and information-precision distributions from observed prices, but whether this is identifiable and computationally tractable at the required scale is untested. Momentum, reversal, and accrual effects \citep{fama_three-factor_1993, jegadeesh_returns_1993} still lack a unified mechanistic account, and whether persona-bot simulations could identify which behavioral primitives generate which anomaly---moving the literature from description toward mechanism---is open. Behavioral feedback loops such as forced selling, margin calls, and herding are also largely absent from regulatory stress tests \citep{basel_basel_2010, dodd_frank_2010}, leaving open what would make a simulation-based stress test regulator-trusted and actionable.

A final cluster concerns fairness, accountability, and governance. Demographic disparities in training data propagate into learned bot behavior and into any policy conclusions drawn from it \citep{shokri_membership_2017, fredrikson_model_2015}, raising the unresolved question of whether the model is meant to describe markets as they are or correct them toward how they should be, and who should decide. Regulatory mandates, academic-industry consortia, and open-source release are competing governance models for who controls access to persona-trained simulators, and which is feasible given the proprietary data such simulators require is not settled. Neural policies are also opaque, and interpretability tools such as LIME, SHAP, and attention remain imperfect, leaving open what auditability standard would be sufficient before a simulation could inform high-stakes policy decisions.

\subsection{Concrete Next Steps}

Moving this agenda forward will require action on several fronts. The most immediate need is for benchmark datasets: public, anonymized, privacy-protected datasets of order-level trading data paired with trader demographics and behavioral covariates would enable reproducible research and lower the barrier to entry for other groups. Open-source simulation frameworks---extending existing ABM platforms such as NetLogo \cite{wilensky_netlogo_1999, tisue_design_2004} or Mesa, or finance-specific libraries such as Zipline, with modular bot-agent components and standard evaluation metrics---would further this goal. Progress is also likely to require cross-disciplinary collaboration among finance researchers, machine-learning practitioners, behavioral economists, and regulatory agencies, plausibly organized through multi-institutional research consortia, alongside direct regulatory engagement: central banks and securities regulators (the SEC, ESMA, the PRA) commissioning research on persona-trained simulations for stress-testing and systemic-risk monitoring would help establish best practices and governance standards. Finally, given the human-behavioral data such simulations depend on, ethical frameworks and governance structures analogous to institutional review boards for human-subject research would need to be adapted to large-scale market simulation.

\subsection{Final Remarks}

Financial markets are complex adaptive systems where macroscopic patterns (prices, volatility, crashes) emerge from microscopic interactions of heterogeneous agents. Persona-trained neural-bot simulation offers a powerful tool for understanding these phenomena, complementing econometric time-series analysis, laboratory experiments, and traditional ABM approaches. By grounding agent behavior in empirical behavioral data and modern machine learning, we can build simulations that are simultaneously realistic, interpretable, and actionable for research and policy.

The framework is not a crystal ball; simulations will sometimes diverge from reality in ways we fail to anticipate. But they offer a laboratory for controlled experiments on market microstructure, behavioral mechanisms, and systemic risks that would be infeasible or unethical in the real world. As markets evolve (high-frequency trading, algorithmic market-making, crypto, central-bank digital currencies), simulation-based tools will become increasingly critical for regulators, investors, and researchers seeking to understand and manage financial stability.

\noindent\textit{AI-assisted writing: A large language model assisted with drafting and editing; the author takes full responsibility. AI is not listed as an author.}
\bibliographystyle{plainnat}
\bibliography{report}

\end{document}